%% file: main.tex
\documentclass[runningheads]{llncs}

\usepackage{eccv}

\usepackage{eccvabbrv}

\usepackage{graphicx}
\usepackage{booktabs}

\usepackage[accsupp]{axessibility}  
\usepackage{amssymb}
\usepackage{algorithm}
\usepackage{algpseudocode}
\usepackage{adjustbox}
\usepackage{graphicx}
\usepackage{multirow}
\usepackage{subcaption}
\usepackage[symbol,para]{footmisc}

\usepackage{hyperref}

\usepackage{orcidlink}

\def\rd{\textcolor{red}}

\def\gr{\textcolor{green}}
\def\bl{\textcolor{blue}}

\begin{document}

\title{M2D2M: Multi-Motion Generation from Text with Discrete Diffusion Models} 

\titlerunning{M2D2M: Multi-Motion  Discrete Diffusion Models}

\author{Seunggeun Chi\inst{1,2}$^{*}$\orcidlink{0000-0001-6965-6938} \and
Hyung-gun Chi\inst{2}$^{*}$\orcidlink{0000-0001-5454-3404}\and
Hengbo Ma$^{\ddag}$\and
Nakul Agarwal\inst{1}\and \\
Faizan Siddiqui\inst{1}\and
Karthik Ramani\inst{2}$^\dag$\orcidlink{0000-0001-8639-5135}\and
Kwonjoon Lee\inst{1}$^\dag$\orcidlink{0000-0002-1433-551X}
}

\authorrunning{S.~Chi et al.}

\institute{\hspace{-0.5em}Honda Research Institute USA \and \hspace{-0.5em}Purdue University \\
\email{\{chi65, chi45\}@purdue.edu},\\
\email{ramani@purdue.edu}, \email{kwonjoon\_lee@honda-ri.com}}

\footnotetext[1]{Co-first authors.} 
\footnotetext[2]{Senior authors.}
\footnotetext[3]{Work done at Honda Research Institute.}

\maketitle
\input{sec/0_abstract}
\input{sec/1_intro}
\input{sec/2_related_works}
\input{sec/3_preliminary}
\input{sec/4_method}
\input{sec/5_experiments}
\input{sec/6_conclusion} 
\newpage
\section*{Acknowledgements}
We acknowledge Feddersen Chair Funds and the US National Science Foundation (FW-HTF 1839971, PFI-TT 2329804) for Dr. Karthik Ramani. Any opinions, findings, and conclusions expressed in this material are those of the authors and do not necessarily reflect the views of the funding agency. We sincerely thank the reviewers for their constructive suggestions. 

%
%
\bibliographystyle{splncs04}
\bibliography{main}
\input{sec/X_suppl} \newpage
\end{document}

%% file: sec/0_abstract.tex
\begin{abstract}
We introduce the \textbf{M}ulti-\textbf{M}otion \textbf{D}iscrete \textbf{D}iffusion \textbf{M}odels (M2D2M), a novel approach for human motion generation from textual descriptions of multiple actions, utilizing the strengths of discrete diffusion models. This approach adeptly addresses the challenge of generating multi-motion sequences, ensuring seamless transitions of motions and coherence across a series of actions. The strength of M2D2M lies in its dynamic transition probability within the discrete diffusion model, which adapts transition probabilities based on the proximity between motion tokens, encouraging mixing between different modes. Complemented by a two-phase sampling strategy that includes independent and joint denoising steps, M2D2M effectively generates long-term, smooth, and contextually coherent human motion sequences, utilizing a model trained for single-motion generation. Extensive experiments demonstrate that M2D2M surpasses current state-of-the-art benchmarks for motion generation from text descriptions, showcasing its efficacy in interpreting language semantics and generating dynamic, realistic motions.
\keywords{Text-to-Motion \and Multi-Motion Generation \and VQ-Diffusion}
\end{abstract}

%% file: sec/1_intro.tex
\section{Introduction} \label{sec:intro}
The generation of human motion is a rapidly advancing field with profound applications in areas such as animation \cite{alexanderson2023listen, ao2023gesturediffuclip}, VR/AR \cite{ipsita2021vrfromx, huang2021adaptutar}, and human-computer interaction \cite{kucherenko2019analyzing, yin2022one}. Particularly, the ability to accurately convert textual descriptions into realistic, fluid human motions is not just a remarkable technical achievement but also a crucial step towards more immersive digital experiences. 

Recent progress in human motion generation has seen a surge in the use of diffusion models \cite{kong2023priority, zhang2023generating, tevet2022human, chen2023executing}. These advancements have been critical in aligning textual descriptions with corresponding human motions. 

\begin{figure}[t]
\centering
\includegraphics[width=.90\textwidth]{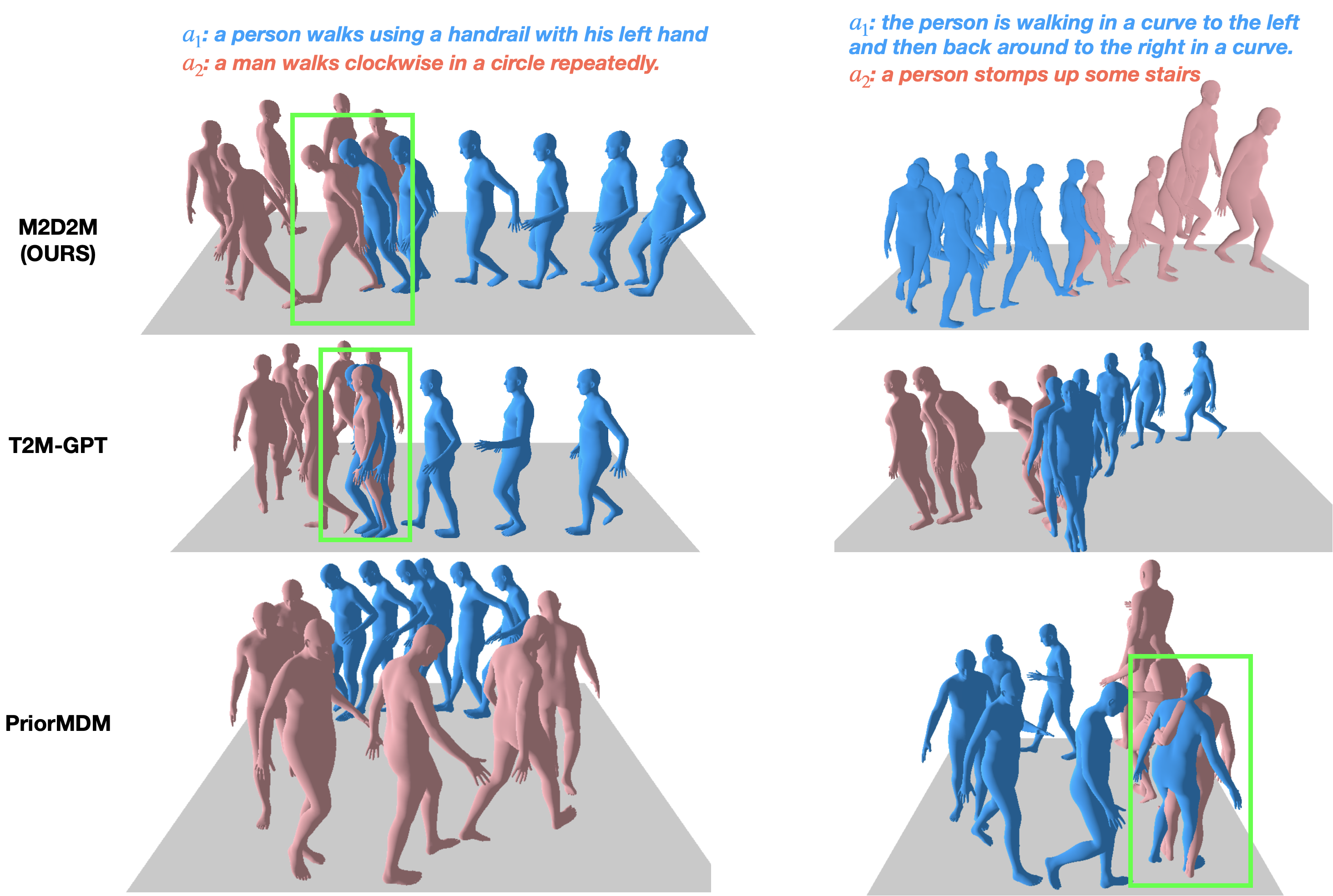}
\caption{\textbf{Qualitative Comparison of Multi-Motion Sequences}. In the transitions highlighted by the \gr{green boxes}, our model shows a consistent and gradual progression of poses compared to others. This indicates that our model not only produces more realistic and smooth motions but also maintains the fidelity of each motion segment, aligning accurately with the corresponding action descriptions on top.
}
\label{fig:qualitative}
\end{figure}

Previous research on human motion generation have mainly focused on single-motion sequences (\textit{i.e.}, sequences that feature a single action), but the ability to generate \textit{multi-motion sequences}, which involve a series of actions, from a set of action descriptions, which is illustrated in \cref{fig:qualitative}, is crucial for many applications. This capability is particularly important in scenarios where a series of actions must be depicted in a continuous and coherent manner, such as in storytelling, interactive gaming, or complex training simulations. However, generating such sequences presents unique challenges where models often struggle to maintain continuity and coherence throughout a series of actions. Previous methods \cite{shafir2023human,zhang2023generating, duan2021single, harvey2020robust, kaufmann2020convolutional, tang2022real}, which generate motion for each action description separately and then attempt to connect them, frequently result in motions with abrupt transitions at action boundaries or distorted individual motions, lacking fidelity to textual descriptions for individual motion.

To tackle this challenge, we introduce the \textbf{M2D2M} (\textbf{M}ulti \textbf{M}otion \textbf{D}iscrete \textbf{D}iffusion \textbf{M}odels), a novel approach for human motion generation from textual descriptions of multiple actions. We devise a novel sampling mechanism to generate coherent and faithful multi-motion sequences using discrete diffusion models \cite{gu2022vector} trained on single-motion sequences. Furthermore, to encourage mixing between different modes (especially at multi-motion boundaries), we introduce a unique transition probability mechanism that considers the proximity between motion tokens.



A key contribution of our work is the introduction of a dynamic transition probability model within the discrete diffusion framework. This model adjusts the transition probabilities based on principles of exploration and exploitation. Initially, it emphasizes broad exploration of diverse motions by choosing elements far apart in the codebook. As the process progresses, the focus shifts to selecting closer elements, refining the probabilities to facilitate convergence towards accurate individual motions, embodying the principle of exploitation.

Our Two-Phase Sampling (TPS) strategy represents another key contribution, addressing the challenge of generating extended human motion sequences. This method starts by sketching a coarse outline of multi-motion sequences through joint sampling, and then it refines each motion with independent sampling. We posit that the exploration of diverse motions is crucial in the initial stage of TPS, where we establish the rough layout of multi-motion sequences. Our ablation results (Table \ref{table:ablation_multi_hml3d}) demonstrate that the synergy between the dynamic transition probability model and TPS is essential for converging to optimal solutions. TPS allows for the generation of multi-motion using models \textit{trained on single-motion generation} without additional training for multi-motion generation, which is particularly advantageous given the \textit{scarcity of datasets containing multiple actions}. TPS enhances the natural flow of the motion, ensuring that transitions between actions are both smooth and realistic.

To quantify the transition behavior of different multi-motion generation models (as seen in \cref{fig:qualitative}), we introduce a novel evaluation metric, \textbf{Jerk}, to measure the smoothness of multi-motion sequences at the transitions between actions. Although similar metrics have been employed in various fields \cite{ldj, jerk2,jerk3,transpose}, to the best of our knowledge, our work is the first to use Jerk for evaluating transition smoothness in multi-motion generation, establishing a specialized benchmark for this task. Experiments show the effectiveness of our approach in enhancing transition smoothness, while maintaining \textbf{fidelity} to \textbf{individual motions} within motion boundaries.


The contribution of our work is summarized as follows:
\vspace{-0.5em}
\begin{itemize}
\item We present a two-phase sampling method for creating multi-motion sequences from text. This method enables multi-motion generation without additional training and offers a better tradeoff between the fidelity of individual motions and the smoothness of transitions between actions \cite{shafir2023human,zhang2023generating}.

\item We introduce a dynamic transition probability for the discrete diffusion model, specifically designed for human motion generation from text.
\item We introduce a new evaluation metric, \textbf{Jerk}, designed to assess the \textit{smoothness} of generated motions at action boundaries.
\item Extensive experiments confirm that our methods establish state-of-the-art performance in both single-motion generation and multi-motion generation.
\end{itemize}

%% file: sec/2_related_works.tex
\section{Related Works} \label{sec:related_works}
\subsection{Human Motion Generation from Text}
Generating 3D human motion from textual descriptions is a growing area within human motion generation \cite{aliakbarian2020stochastic, komura2017recurrent, yan2018mt, yan2019convolutional, mao2020history, mao2019learning, ahn2018text2action, ahuja2019language2pose, ghosh2021synthesis, zhong2023attt2m, zhang2023remodiffuse, zhang2024finemogen}. Recent advancements include using CLIP \cite{radford2021learning} for text encoding and aligning text and motion embeddings \cite{tevet2022motionclip,petrovich2022temos} alongside the development of motion-language datasets \cite{plappert2016kit, punnakkal2021babel, guo2022generating}. Notably, diffusion-based models \cite{zhang2022motiondiffuse, tevet2022human, chen2023executing, zhang2023remodiffuse, zhang2024finemogen} have been increasingly used for text-to-motion generation.

\begin{figure*}[t!]
\centering
\includegraphics[width=1.0\linewidth]{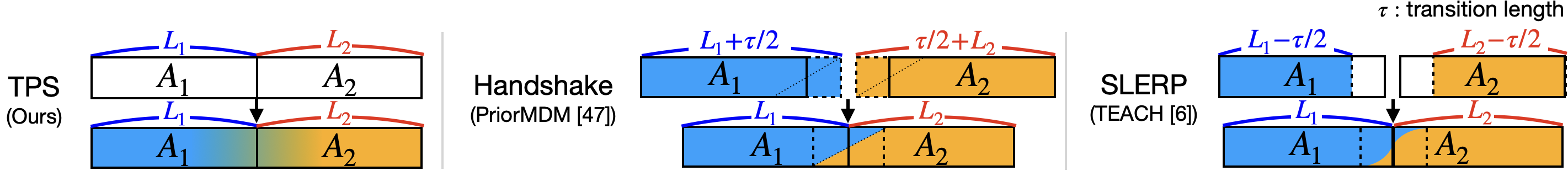}
\caption{\textbf{Comparison of Multi-Motion Generation Algorithms.} Unlike heuristic post-processing methods for combining independent motions such as Handshake \cite{shafir2023human} and SLERP \cite{TEACH:3DV:2022}, TPS is a single-stage algorithm for a multi-motion generation that does not require completed individual motions or a hyper-parameter for transition length.}
\label{fig:algorithm}
\end{figure*}

Multi-motion generation is crucial for creating realistic and continuous human motion sequences. Recent work, such as ``motion in-betweening" \cite{duan2021single, harvey2018recurrent, harvey2020robust, kaufmann2020convolutional, tang2022real}, addresses this by interpolating between motions. TEACH \cite{TEACH:3DV:2022} enhanced this approach with an unfolding method using SLERP interpolation. Additionally, PriorMDM \cite{shafir2023human} introduces a handshake algorithm for smoother transitions in long sequences. Despite their effectiveness in bridging gaps, these methods require an additional stage for multi-motion generation to merge independently generated motions. As illustrated in \cref{fig:algorithm}, these methods not only modify the length of individual motions but also require an extra hyper-parameter for transition length to achieve smoothness, affecting the outcomes and evaluation metrics. VAE-based methods like Multi-Act \cite{Lee2023MultiAct} and TEACH \cite{TEACH:3DV:2022} attempt to generate motion conditioned on a previous motion and a text but face limitations in long-term generation due to their iterative process. FineMoGen \cite{zhang2024finemogen} utilzes spatial-temporal attention for fine-grained text description to generate multi-motion. Our method overcomes these challenges by maintaining fidelity and smoothness in the generated motion, significantly enhancing multi-motion generation without the trade-offs inherent in traditional approaches.

\begin{figure*}[t!]
\centering
\includegraphics[width=1.0\linewidth]{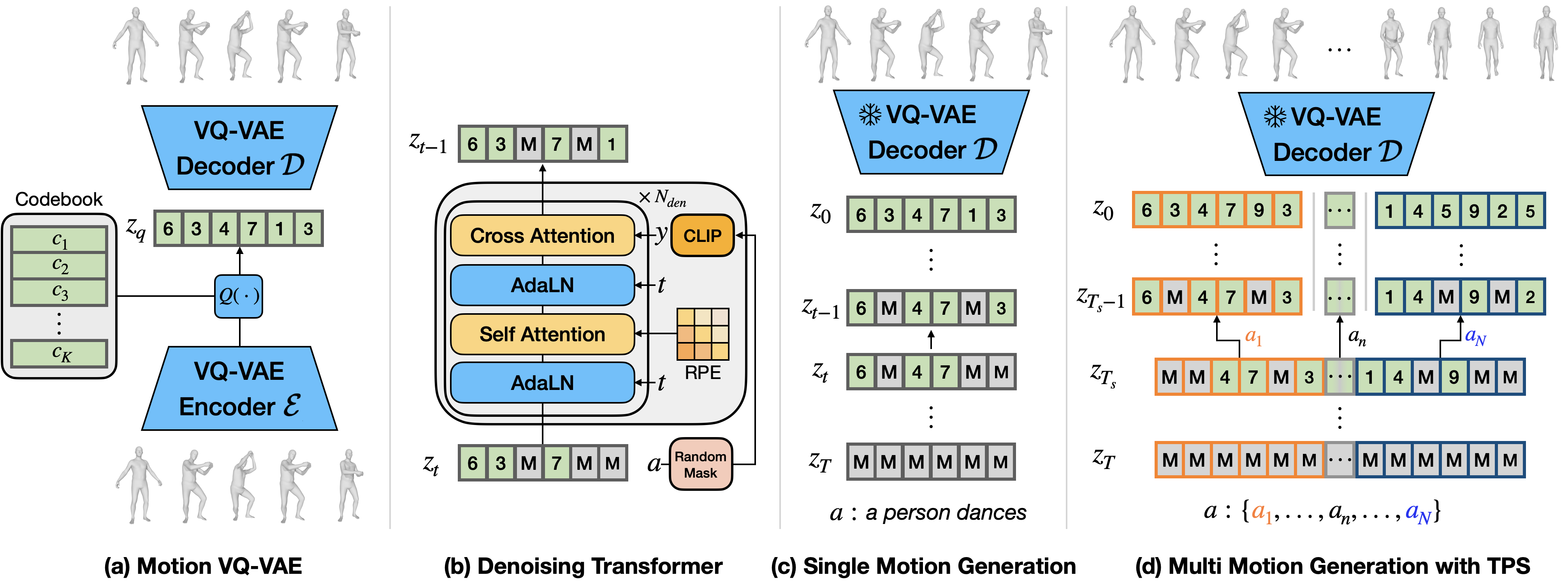}
\caption{\textbf{Overview of M2D2M}. We train a \textbf{(a)} VQ-VAE to obtain motion tokens, which is subsequently used to train a \textbf{(b)} Denoising Transformer for the discrete diffusion model. In generating human motion, we follow the \textbf{(c)} standard denoising process for single-motion generation and \textbf{(d)} employ Two-Phase Sampling (TPS) for multi-motion generation. A \texttt{<MASK>} token is denoted as `\textbf{M}' in the figure.}
\label{fig:Architecture}
\end{figure*}

\subsection{Discrete Diffusion Models}
Diffusion models \cite{sohl2015deep}, defined by forward and reverse Markov processes, are integral to advancements in generative models. These models, known for transforming data into increasingly noisy variables and subsequently denoising them, benefit from stability and rapid sampling capabilities. Enhanced by neural networks learning the reverse process \cite{dhariwal2021diffusion, ho2020denoising, song2020improved, gu2022vector}, they are particularly effective in continuous spaces like images. Latent diffusion models \cite{rombach2022high}, operating in a latent space before returning to the original data space, adeptly handle complex data distributions.

In discrete spaces such as text, diffusion models also excel. D3PM \cite{austin2021structured} and VQ-Diffusion \cite{gu2022vector} have introduced methods like structured categorical corruption and mask-and-replace to minimize errors in iterative models. This showcases the broad applicability of diffusion models. Drawing inspiration from recent studies demonstrating VQ-VAE's \cite{van2017neural} effectiveness in human pose modeling in discrete spaces \cite{kong2023priority, zhang2023generating}, we apply discrete diffusion to human motion generation. Similar to our approach Kong et al \cite{kong2023priority} also introduce discrete diffusion for human motion generation from the text. However, different from this work, we introduce a new transition matrix considering the relationship between action tokens and is specifically tailored for multi-motion generation.

%% file: sec/3_preliminary.tex
\section{Preliminary: Discrete Diffusion Models}\label{sec:prelim}
Discrete diffusion models \cite{hoogeboom2021argmaxflowsmultinomialdiffusion, austin2021structured, gu2022vector} are a class of diffusion models which work by gradually adding noise to data and learning to reverse this process. Unlike continuous models like latent diffusion models \cite{rombach2022high} which operate on data represented in a continuous space, discrete diffusion models work with data representation in discrete state spaces.

\noindent\textbf{Forward Diffusion Process.}
Since the first introduction of the discrete diffusion model \cite{sohl2015deep}, VQ-Diffusion \cite{gu2022vector} has improved the approach by incorporating a mask-and-replace strategy.
VQ-Diffusion entails a forward diffusion process by transitioning from one token to another token. 
The forward Markov diffusion process for step $t\!-\!1$ to $t$ is given by:
\begin{flalign}
q(z_{t}|z_{t-1}) = \boldsymbol{v}^{\top}(z_t)\mathbf{Q}_t\boldsymbol{v}(z_{t-1}),
\end{flalign}
where $ \boldsymbol{v}(z_t) \in \mathbb{R}^{(K+1) \times 1}$ denotes the one-hot encoded vector for the token index of $z_t$, and $\mathbf{Q}_t[i,j]$ is the transition probability from a token $z_i$ to $z_j$ at diffusion step $t$. The transition probability matrix $\mathbf{Q}_t \in \mathbb{R}^{(K+1) \times (K+1)}$ is written as:
\begin{flalign}
\mathbf{Q}_{t} =
\setlength\arraycolsep{0.7em}
\left[
\begin{array}{c|c}
  \hat{\mathbf{Q}}_{t} & 0 \\
  \hline
  \gamma_{t}\cdot\mathbf{1}^\top & 1
\end{array}
\right], \text{ where } \hat{\mathbf{Q}}_{t} = \alpha_{t}\mathbf{I}+\beta_{t}\mathbf{1}\mathbf{1}^\top.
\label{eq:q_t}
\end{flalign}
Here, $\mathbf{I}$ is the identity matrix, $\mathbf{1}$ is a column vector of ones, $\beta_t$ represents the probability of transitioning between the different tokens, $\gamma_t$ denotes the probability of transitioning to a \texttt{<MASK>} token, and $\alpha_t\!=\!1\!-\!K\beta_t\!-\!\gamma_t$.
Due to the Markov property, the probabilities of $z_t$ at arbitrary diffusion time step can be derived $q(z_{t}|z_{0}) = \boldsymbol{v}^{\top}(z_t)\overline{\mathbf{Q}}_t\boldsymbol{v}(z_{0})$, where $\overline{\mathbf{Q}}_t = \mathbf{Q}_t \mathbf{Q}_{t-1}\cdots \mathbf{Q}_1$.
The matrix is constructed such that the \texttt{<MASK>} token always maintains its original state so that $z_t$ converges to \texttt{<MASK>} token with sufficiently large $t$.

\noindent\textbf{Conditional Denoising Process.}
The conditional denoising process through a neural network $p_{\theta}$. This network predicts the noiseless token $z_0$ when provided with a corrupted token and its corresponding condition, such as a language token. 
For training the network $p_\theta$, beyond the denoising objective, the training incorporates the standard variational lower bound objective \cite{sohl2015deep}, denoted as $\mathcal{L}_{\text{vlb}}$. The training objective with a coefficient for the denoising loss $\lambda$ is:
\begin{flalign}
\mathcal{L} = \mathcal{L}_{\text{vlb}} + \lambda \mathbb{E}_{z_t \sim q(z_t|z_0)}[-\log{p_{\theta}(z_0|z_t,y)}],
\label{eq:vq_diffusion}
\end{flalign}
Here, the reverse transition distribution can be written as follow:
\begin{flalign}
p_{\theta}(z_{t-1} | z_t, y) = {\textstyle\sum}^{K}_{\tilde{z}_0=1} q(z_{t-1} | z_t, \tilde{z}_0) p_{\theta}(\tilde{z}_0 | z_t, y).
\end{flalign}
By iteratively denoising tokens from $T$ down to 1, we can obtain the generated token $z_0$ conditioned on $y$.
The tractable posterior distribution of discrete diffusion can be expressed as:
\begin{flalign}
&q(z_{t-1}|z_t,z_0) = \frac{q(z_t|z_{t-1},z_0)q(z_{t-1}|z_0)}{q(z_t|z_0)} \nonumber \\
&= \frac{\big(\boldsymbol{v}^{\top}(z_t)\mathbf{Q}_t\boldsymbol{v}(z_{t-1})\big)\big(\boldsymbol{v}^{\top}(z_{t-1})\mathbf{\overline{Q}}_{t-1}\boldsymbol{v}(z_{0})\big)}{\boldsymbol{v}^{\top}(z_t)\boldsymbol{\mathbf{Q}}_t\boldsymbol{v}(z_{0})}.
\end{flalign}

%% file: sec/4_method.tex
\section{Multi-Motion Discrete Diffusion Model} \label{sec:method}
We introduce the \textbf{M2D2M} (\textbf{M}ulti \textbf{M}otion \textbf{D}iscrete \textbf{D}iffusion \textbf{M}odel), as illustrated in \cref{fig:Architecture}. M2D2M is a discrete diffusion model designed specifically for generating human motion from textual descriptions of multiple actions. This model utilizes a VQ-VAE-based discrete encoding \cite{van2017neural} which has proven effective in representing human motion \cite{kong2023priority, zhang2023generating, jiang2023motiongpt}. The following sections will provide a detailed overview of our approach.

\subsection{Motion VQ-VAE}
To establish a codebook for discrete diffusion, we first trained a VQ-VAE model \cite{van2017neural}. Our training approach and the model's architectural design closely follow \cite{zhang2023generating}. The model comprises of an encoder $\mathcal{E}(\cdot)$, a decoder $\mathcal{D}(\cdot)$, and a quantizer $\mathcal{Q}(\cdot)$ (see \cref{fig:Architecture} \textbf{(a)}). The encoder processes human motion, represented by $\mathbf{x} \!\in\! \mathbb{R}^{L \times D}$, converting it into motion tokens, $\mathbf{z}\!=\!\mathcal{E}(\mathbf{x})\!\in\! \mathbb{R}^{\frac{L}{4} \times D}$. Here, $L$ signifies the length of the motion sequence and $D$ denotes the dimensionality of each codebook. The quantizer's role is to map the motion token at any timeframe $\tau$ to the nearest codebook entry, determined by $\mathbf{z}_q[\tau]\!=\!\mathcal{Q}(\mathbf{z}[\tau])\!=\!\text{argmin}_{c_i \in \mathcal{C}}{||\mathbf{z}[\tau] - c_i||_2}$. Here, $\mathcal{C}\!=\!\{c_1, \ldots, c_K\}$ represents the codebook, where $K$ signifies the total number of codebooks. Subsequently, the decoder utilizes these motion tokens to reconstruct the human motion as $\hat{\mathbf{x}}\!=\!\mathcal{D}(\mathbf{z})$. We train the motion VQ-VAE with
\begin{equation}
\mathcal{L}_{\text{VQ}}=||\mathbf{x}\!-\!\hat{\mathbf{x}}||_2 \!+\! ||\mathbf{z}_q\!-\!\text{sg}[{\mathbf{z}}]||_2 \!+\! \lambda_{\text{VQ}} ||\text{sg}[\mathbf{z}_q]\!-\!\mathbf{z}||_2,
\end{equation}
where $\text{sg}[\cdot]$ represents stop gradient and $\lambda_{\text{VQ}}$ is coefficient for commitment loss.

\subsection{Dynamic Transition Probability} \label{sec:transition}
\begin{figure}[!t]
\centering
\begin{subfigure}[h]{0.45\linewidth}
\vspace{-1.5em}
\includegraphics[width=0.9\linewidth]{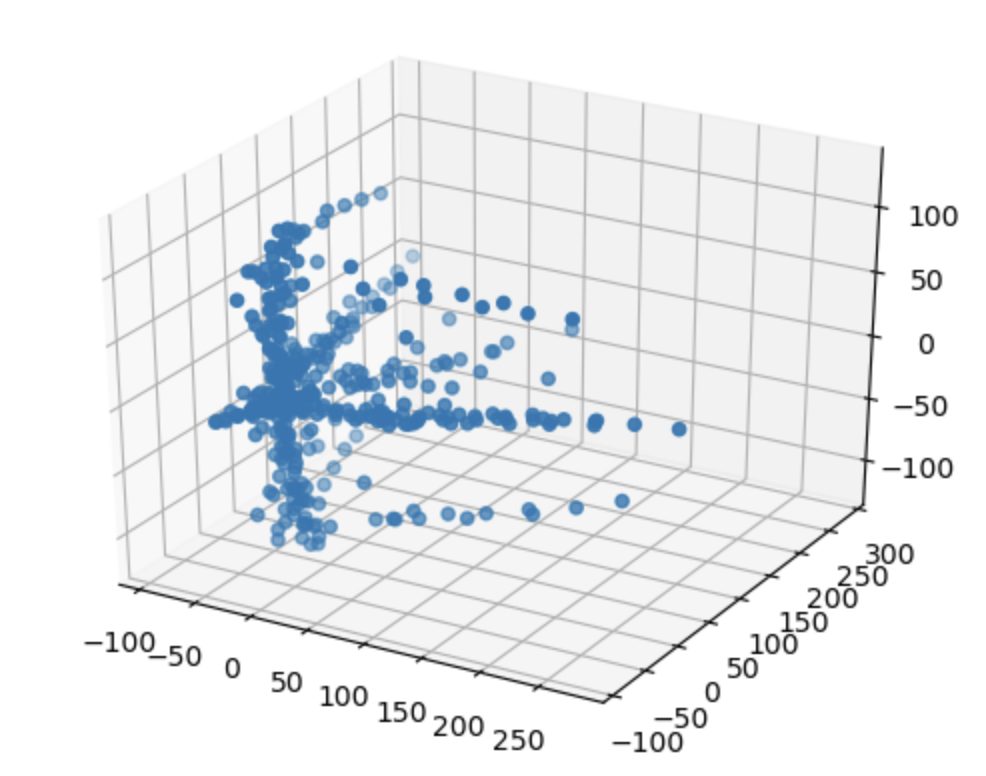}
\caption{Latent Codebook Representation}
\end{subfigure}
\begin{subfigure}[h]{0.45\linewidth}
\includegraphics[width=1.0\linewidth]{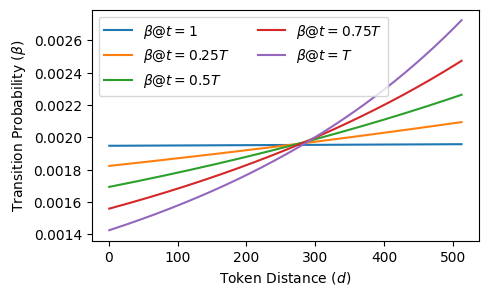}
\caption{Transition Probability}
\end{subfigure}
\vspace{-0.5em}
\caption{(a) PCA plot representing motion tokens from the codebook of Motion VQ-VAE visualized in 3D space. (b) Plot of the dynamic transition probability function $\beta(t,d)$ across various diffusion steps $t$.}
\vspace{-1.2em}
\label{fig:beta}
\end{figure}

In the VQ-Diffusion model \cite{gu2022vector}, the transition matrix $\mathbf{Q}_t$ employs a uniform transition probability $\beta_t$ across different tokens, as described in \cref{eq:q_t}. This method overlooks the varying proximity between motion tokens, which is essential for capturing the context of human motion. To address this, we introduce a dynamic transition probability that accounts for the distance between tokens. During the initial stages of diffusion, when the diffusion step $t$ is large, our model adopts an exploratory approach, allowing for a wide range of transitions to foster diversity. As $t$ progresses towards $0$, the model gradually reduces the favorability of transitions between more distantly related tokens, eventually becoming uniform, identical to the original VQ-Diffusion model \cite{gu2022vector}, as shown in \cref{fig:beta}.  The strategy is to begin with broad exploration and progressively narrow the focus as diffusion steps decrease, thereby improving the precision and coherence in generating extended motion sequences.

Our approach mathematically formulates the transition probability at each diffusion step $t$ as $\beta(t, d)$, where $d$ signifies the distance between codebook tokens. The transition probability is defined by the following equation:
\begin{flalign}
\beta(t, d) = (1 - \gamma_t - \alpha_t) \cdot \text{softmax}_d\left(\eta\cdot\frac{t}{T} \cdot \frac{d}{K} \right),
\end{flalign}
where $\eta$ is a scale factor that modulates the influence of the softmax function on the relative distances between tokens.
The essence of this equation lies in its softmax function over distances, which progressively assigns higher probabilities to greater distances between tokens as the diffusion step $t$ advances. This allocation adheres to the transition probability constraint $\gamma_t\!+\!\alpha_t\!+\!\sum_{d=1}^{K}\beta(t,d)\!=\!1$. The distance-based modulation, scaled by $\eta\cdot\frac{t}{T}\cdot\frac{d}{K}$, ensures that as the diffusion process unfolds, the selection of token transitions is increasingly influenced by the distance metric. This approach boosts early stage exploration of denoising by favoring transitions to distant tokens. The transition matrix is defined as
\begin{flalign}
\mathbf{Q}_{t} =
\setlength\arraycolsep{0.7em}
\left[
\begin{array}{c|c}
  \hat{\mathbf{Q}}_{t} & 0 \\
  \hline
  \gamma_{t}\cdot\mathbf{1}^\top & 1
\end{array}
\right], \text{ where } \hat{\mathbf{Q}}_{t} = \alpha_{t}\mathbf{I}+\beta_{(t,d_{i,j})}\mathbf{1}\mathbf{1}^\top.
\label{eq:q_t}
\end{flalign}
In this matrix, each element \(d_{i,j}\) represents the distance \(d(z_i, z_j)\) for indices \(i,j\) ranging from 1 to \(K\), where \(i\) and \(j\) denote the row and column indices, respectively. Here, \(d(\cdot, \cdot)\) is a distance metric specifically chosen as the rank index of codebook entries, which are sorted by their L2 distances. The dynamic and context-sensitive nature of this matrix formulation allows for an adaptive approach to the diffusion process, modifying transition probabilities in response to the evolving state of the diffusion and the relative distances between tokens.

\begin{algorithm}[t]
\caption{Two-Phase Sampling (TPS).}
\label{alg:sampling}
\textbf{Given}: Action sentences $\mathbf{a} = \{a^1, \ldots, a^N\}$\\
\textbf{Hyperparameter}: $T_s$
\begin{algorithmic}[1] 
\State $\mathbf{z}_T \sim p(\mathbf{z}_T)$
\State $\mathbf{y} \leftarrow \text{CLIP-TextEncoder}(\mathbf{a})$

\For{$t = T, T\!-\!1, \ldots, T_s\!+\!1$} \Comment{Joint Sampling}
\State $\mathbf{z}_{t-1} \sim p_{\theta}(\mathbf{z}_{t-1}|\mathbf{z}_{t}, \mathbf{y})$
\EndFor

\For{$t = T_s, T_s\!-\!1, \ldots, 1$} \Comment{Independent Sampling}
\For{$i = 1, \ldots, N$} \Comment{Parallel Process}
 \State $\mathbf{z}_{t-1}^i \sim p_{\theta}(\mathbf{z}_{t-1}^i|\mathbf{z}_{t}^i, y^i)$
\EndFor
\EndFor
\State $\mathbf{z}_{0} \leftarrow \text{Concat}(\{\mathbf{z}_{0}^1,\ldots,\mathbf{z}_{0}^N\})$
\State $\mathbf{\hat{x}} = \mathcal{D}(\mathbf{z}_0)$

\State \Return $\mathbf{\hat{x}}$
\end{algorithmic}
\label{alg:sampling}
\end{algorithm}
\subsection{Sampling for Multi-Motion Generation} \label{sec:sampling}
We introduce a Two-Phase Sampling (TPS) method for the discrete diffusion model, designed to generate long-term human motion sequences from a series of action descriptions $\mathbf{a} = {a^1, \ldots, a^N}$. This approach enables the creation of multi-motion sequences using a model trained for single-motion generation. The overview of TPS is presented in \cref{alg:sampling} and visually depicted in \cref{fig:Architecture} \textbf{(d)}. In the algorithm, subscripts represent diffusion steps, while superscripts denote action indices. TPS effectively overcomes the challenge of ensuring smooth transitions between distinct actions, while preserving the distinctiveness of each motion segment as per its action description.

The denoising process begins by outlining the basic contours of the entire action sequence, subsequently refining these outlines to achieve semantic coherence with the textual descriptions. 
Inspired by this approach, our two-phase sampling starts with \textit{joint sampling}, where mask tokens from different actions are merged and collectively denoised using a denoising Transformer. This allows self-attention mechanism within the Transformer to integrate contextual information from action descriptions, ensuring that tokens influence each other to achieve seamless motion transitions. This step is followed by \textit{independent sampling}, wherein each action is individually denoised within its designated boundaries to align accurately with its specific description.

The number of joint denoising steps, denoted by $T_s$, is carefully adjusted to achieve smooth transitions without losing the distinctiveness of each action. 

\subsection{Denoising Transformer}
Motivated by the work of VQ-Diffusion \cite{gu2022vector}, we design a denoising transformer that estimates the distribution $p_{\theta}(\tilde{z}_0|z_t, y)$ using the Transformer architecture \cite{vaswani2017attention}. An overview of our proposed model is depicted in \cref{fig:Architecture} \textbf{(b)}.  To incorporate the diffusion step $t$ into the network, we employ the adaptive layer normalization (AdaLN) \cite{ba2016layer,lee2022vitgan}. The action sentence $a$ is encoded into the action token $y$ using the CLIP \cite{radford2021learning} text encoder. The Transformer's cross-attention mechanism then integrates this action information with motion, providing a nuanced conditioning with the action sentence. To enhance the human motion generation of our Transformer architecture, we added the following features:

\noindent\textbf{Relative Positional Encoding.} One of our primary objectives is the generation of long-term motion sequences. During the training phase, models exclusively trained on single-motion struggle to generate longer sequences. This limitation is observed when using traditional absolute positional encodings \cite{vaswani2017attention} that assign a static vector to each position, confining the model's capability to the maximum sequence length encountered during its training. By leveraging Relative Positional Encoding (RPE) \cite{raffel2020exploring}, we equip our models with the ability to extrapolate beyond the sequence lengths experienced in training, thus significantly enhancing their proficiency in generating extended motion sequences.

\noindent\textbf{Classifier-Free Guidance.}
We adopt classifier-free guidance \cite{ho2022classifier}. This approach facilitates a balance between diversity and fidelity, allowing both conditional and unconditional sampling from the same model. For unconditional sampling, a learnable null token, denoted as $\varnothing$, may be substituted for the action token $y$. The action token $y$ is replaced by $\varnothing$ with a probability of 10\%. When inference, the denoising step is defined using $s$ as follows: 
\begin{flalign}
\log p_\theta(z_{t-1} | z_t, y) = (s + 1)&\log p_\theta(z_{t-1} | z_t, y) - s\log p_\theta(z_{t-1} | z_t, \varnothing).
\label{eq:classifier_free}
\end{flalign}

%% file: sec/5_experiments.tex
\section{Experiments} \label{sec:experiment}
Our experiments are designed to assess the capabilities of our model on two tasks: 1) multi-motion generation (\cref{sec:multi_motion_exp}) and 2) single-motion generation (\cref{sec:single_motion_exp}). These experiments are conducted using the following motion-language datasets.
\noindent\textbf{HumanML3D} \cite{Guo_2022_CVPR} is the largest dataset in the domain of language-annotated 3D human motion, boasting 14,616 sequences of human motion, each meticulously aligned to a standard human skeleton template and recorded at 20 FPS. Accompanying these sequences are 44,970 textual descriptions, with an average length of 12 words. Notably, each motion sequence is associated with a minimum of three descriptive texts.
\noindent\textbf{KIT-ML} \cite{plappert2016kit} comprises 3,911 human motion sequences, each annotated with one to four natural language descriptions, culminating in a total of 6,278 descriptions, averaging 8 words in length. The dataset collated motion sequences sourced from both the KIT \cite{mandery2015kit} and CMU \cite{cmumocap} datasets, with each sequence downsampled to 12.5 FPS. Each motion clip in this collection is paired with one to four corresponding textual descriptions.

In our experiments, we perform 10 evaluations for each model, providing a comprehensive analysis of their performance. We report both the mean and standard deviation of these evaluations, along with a 95\% confidence interval to ensure statistical robustness. Detailed information about our model implementation can be found in Appendix \textcolor{red}{A}.

\subsection{Evaluation of Generated Motions}
\noindent\textbf{Single-Motions.} To evaluate generated single-motions, we adopt evaluation metrics from previous works \cite{guo2022generating, kong2023priority}. 1) \textbf{R-Top3} measures the model's precision in correlating motion sequences with their corresponding textual descriptions, highlighting the importance of accurate retrieval from a range of options. 2) \textbf{FID} (Frechet Inception Distance) evaluates the realism of generated motions by comparing the distribution of generated data with real data. 3) \textbf{MM-Dist} (Multi-Modal Distance) gauges the average closeness between features of generated motions and the text features of their respective descriptions, ensuring effective synchronization across different modalities. 4) \textbf{Diversity} assesses the range of generated motions, reflecting the natural variation in human movement. 5) \textbf{MModality} (Multi-Modality) examines the model's capacity to produce a wide array of plausible motions from a single text prompt, a key aspect for versatile motion generation.

\begin{figure}[!t]
\centering
\begin{subfigure}[h]{0.4\linewidth}
\includegraphics[width=1.0\linewidth]{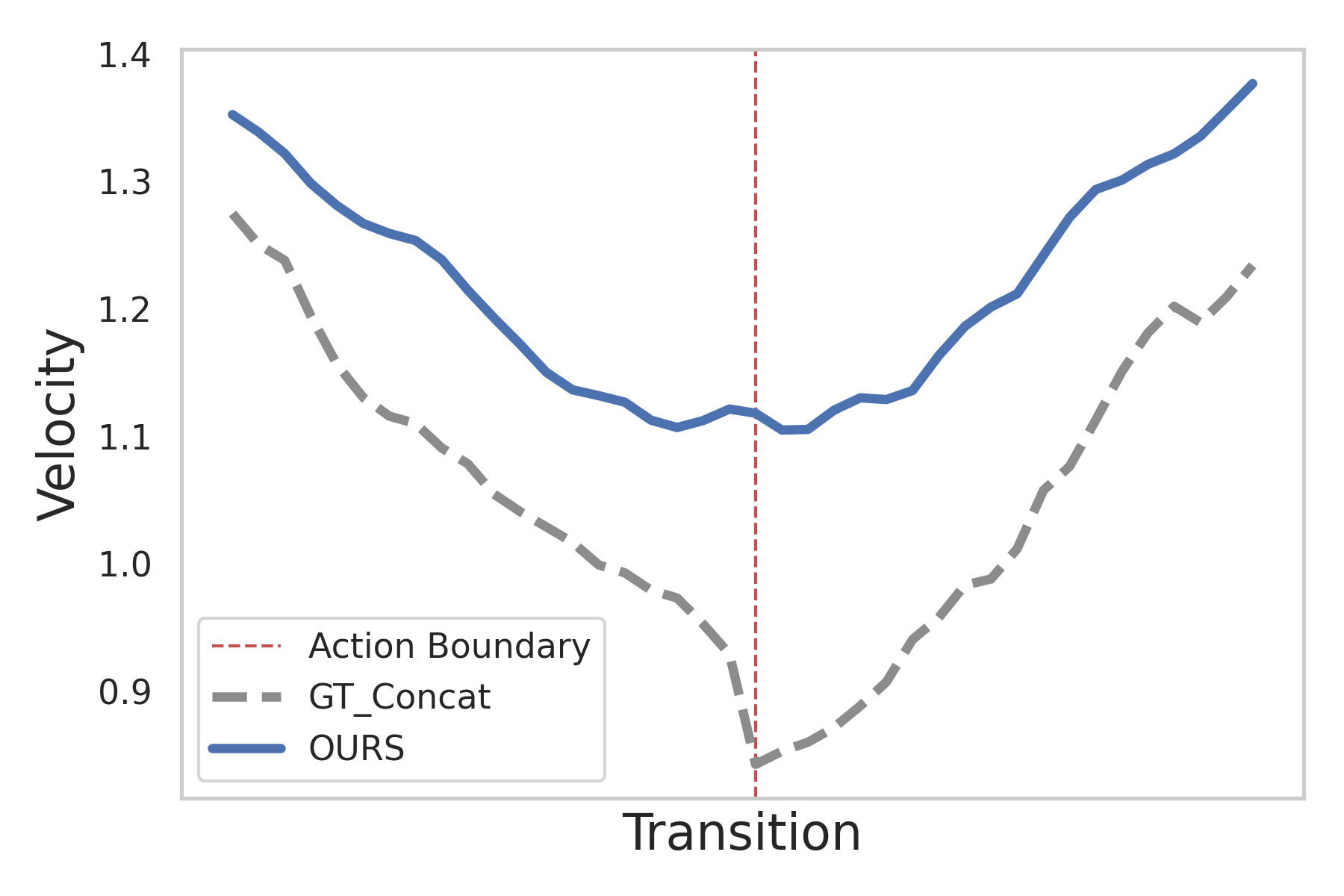}
\caption{Velocity}
\end{subfigure}
\begin{subfigure}[h]{0.4\linewidth}
\includegraphics[width=1.0\linewidth]{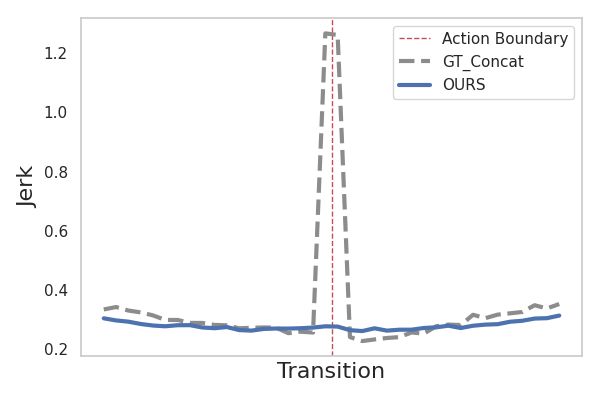}
\caption{Jerk}
\end{subfigure}
\caption{Plots for \textbf{average motion transition} produced by our approach versus a concatenation of two randomly selected real motions: \textbf{(a)} Velocity. \textbf{(b)} Jerk (time derivative of acceleration normalized by peak velocity).}
\label{fig:smooth_fourier}
\end{figure}

\noindent\textbf{Multi-Motions.} 
To evaluate the fidelity of each individual motion within our generated multi-motion sequences, we compute key metrics used to evaluate single-motion: \textbf{R-Top3}, \textbf{FID}, \textbf{MM-Dist}, and  \textbf{Diversity}. This is measured by segmenting the generated multi-motion into distinct parts, each corresponding to a specific action description. Through this process, we can precisely analyze the alignment between each motion segment and its corresponding action. To evaluate the transition part for multi-motion generation, \textbf{FID} and \textbf{Diversity} are evaluated for 40 frames of motion around boundaries, following the methodology explained by \cite{shafir2023human}. Additionally, we defined a metric for evaluating the smoothness of multi-motion generation near motion boundaries.  
\textbf{Jerk} \cite{ldj, jerk2, jerk3, flash1985coordination} is based on the derivative of acceleration with respect to time. Similar to \cite{jerk2}, we use the logarithm of a dimensionless jerk to achieve scale invariance with respect to velocity. Contrary to the \textit{Jitter} described in the prior study \cite{yi2021transpose}, which measures the smoothness of individual motions, we calculate the integral of the time derivative for each joint in a transitional motion. We then average these values to derive a single statistic that represents the smoothness of the transition.
\begin{flalign}
Jerk = \sum_{p}\ln\frac{1}{v^2_{p,\text{peak}}} \int_{t_1}^{t_2} \lVert \frac{d}{dt}\mathbf{a}_p(t) \rVert_2 ^ 2 \,dt,
\label{eq:jerk1}
\end{flalign}
where $p$ denotes each joint, $v_{p,\text{peak}}$ is the maximum speed of the joint $p$ and $[t_1, t_2]$ is the time interval for the motion transition. The metric's intuitive significance becomes clear when comparing a \textbf{single-motion transition} generated by our method to the concatenation of two arbitrarily chosen real motions. As depicted in \cref{fig:smooth_fourier}, directly joining different real motions results in considerable jerk. Conversely, our method, which integrates two motions seamlessly using a two-phase sampling strategy, greatly diminishes jerk. This method facilitates a smooth transition between motions, resulting in minimal jerk.

\subsection{Multi-Motion Generation} \label{sec:multi_motion_exp} 
The objective of multi-motion generation is to produce continuous human motion sequences from a series of action descriptions. To achieve this, we adapt our model, initially trained for single-motion generation, to handle extended sequences. This adaptation is facilitated by our two-phase sampling strategy, as elaborated in \cref{sec:sampling}. To assess the model's effectiveness in multi-motion generation, we created a test set by randomly combining $N$ action sentences in the test set. This process yielded a total of 1,448 test instances for the HumanML3D and 532 instances for the KIT-ML for $N\!=\!4$ scenario. We provide further details about test set generation for multi-motion generation in Appendix \textcolor{red}{B}. This approach ensures a thorough evaluation of the model's capability to generate coherent long-term motion sequences from multiple action descriptions.

\noindent\textbf{Baselines.}
We selected PriorMDM \cite{shafir2023human} and T2M-GPT \cite{zhang2023generating} as baseline models for the multi-motion generation task. For the PriorMDM model, we adhered to the original configuration without making any modifications. For the T2M-GPT, we extend the approach by concatenating the codebook for each motion and feeding it into the decoder to accommodate the auto-regressive nature of GPT models. Implementation details of these models can be found in Appendix \textcolor{red}{A}. Importantly, we opted not to include iterative multi-motion generation models such as TEACH \cite{TEACH:3DV:2022} and Multi-Act \cite{Lee2023MultiAct} in our comparison. This decision was based on the fact that these models rely on the relationships between subsequent actions, which are not represented in the HumanML3D and KIT-ML datasets and would require additional annotations for proper evaluation.


\begin{table}[!t]
\centering
\caption{Multi-motion generation performance on HumanML3D. `Individual Motion' denotes individual motions within our motion boundaries. For transitions, ground truth (single) motions are independently sampled to match the transition length, while ground truth (concat) involves concatenating ground truth motions sampled from the same textual condition with generated motions.}
\resizebox{1.0\linewidth}{!}{
    \addtolength{\tabcolsep}{4pt}
    \begin{tabular}{c|cccc|ccc}
        \hline
        \multirow{2}{*}{Methods} & \multicolumn{4}{c}{Individual Motion} & \multicolumn{3}{c}{Transition (40 frames)}\\\cline{2-5}\cline{5-8}
         & R-Top3$\uparrow$ & FID$\downarrow$ & MMdist$\downarrow$ & Div$\rightarrow$ & FID$\downarrow$ & Div$\rightarrow$ & Jerk$\rightarrow$ \\
        \hline\hline
        Ground Truth (Single) & $0.791^{\pm.002}$ & $0.002^{\pm.000}$ & $2.707^{\pm.008}$ & $9.820^{\pm.065}$ & $0.003^{\pm.002}$ & $9.574^{\pm.054}$ & $1.192^{\pm.005}$ \\
        Ground Truth (Concat) & - & - & - & - & - & - & $1.371^{\pm.004}$ \\\hline
        PriorMDM \cite{shafir2023human}         & $0.586^{\pm.003}$ & $0.832^{\pm.017}$ & $5.901^{\pm.021}$  & $9.543^{\pm.005}$ & $3.351^{\pm.034}$ & $\textbf{8.801}^{\pm.098}$ & $0.476^{\pm.004}$  \\
        T2M-GPT \cite{zhang2023generating}      & $0.719^{\pm.003}$ & $0.342^{\pm.019}$ & $ 3.512^{\pm.014} $ & $9.692^{\pm.003}$ & $3.412^{\pm.027}$ & $8.716^{\pm.135}$ & $1.321^{\pm.005}$\\\hline
        \textbf{M2D2M}                 & $\textbf{0.733}^{\pm.003}$ & $\textbf{0.253}^{\pm.016}$ & $\textbf{3.165}^{\pm.019} $ & $\textbf{9.806}^{\pm.005}$ & $\textbf{3.276}^{\pm.024}$ & $8.599^{\pm.154}$ & $\textbf{1.238}^{\pm.008}$ \\\hline
    \end{tabular}
    }
\label{table:longterm_hml3d}
\end{table}

\begin{table}[!t]
\centering
\caption{Multi-motion generation performance on KIT-ML.}
\resizebox{1.0\linewidth}{!}{
    \addtolength{\tabcolsep}{4pt}
    \begin{tabular}{c|cccc|ccc}
        \hline
        \multirow{2}{*}{Methods} & \multicolumn{4}{c}{Individual Motion} & \multicolumn{3}{c}{Transition (40 frames)}\\\cline{2-5}\cline{5-8}
         & R-Top3$\uparrow$ & FID$\downarrow$ & MMdist$\downarrow$ & Div$\rightarrow$ & FID$\downarrow$ & Div$\rightarrow$ & Jerk$\rightarrow$  \\
        \hline\hline
        Ground Truth (Single) & $0.775^{\pm.008}$ & $0.034^{\pm.004}$ & $2.779^{\pm.019}$ & $11.055^{\pm.122}$ & $0.041^{\pm.005}$ & $10.434^{\pm.044}$ & $1.231^{\pm.002}$ \\
        Ground Truth (Concat) & - & - & - & - & - & - & $1.469^{\pm.003}$ \\\hline
        PriorMDM \cite{shafir2023human}             & $0.292^{\pm.217}$ & $3.311^{\pm.106}$ & ${5.451}^{\pm.045}$ & $\textbf{10.842}^{\pm.067}$ & $21.231^{\pm.844}$ & $\textbf{7.281}^{\pm.045}$ & $0.594^{\pm.002}$\\
        T2M-GPT \cite{zhang2023generating}          & $0.667^{\pm.006}$ & $0.907^{\pm.059}$ & $3.421^{\pm.026}$  & ${10.587}^{\pm.089}$ & $\textbf{14.494}^{\pm.547}$ & $7.059^{\pm.042}$ & $1.388^{\pm.003}$  \\\hline
        \textbf{M2D2M}  & $\textbf{0.711}^{\pm.006}$ & $\textbf{0.817}^{\pm.058}$ & $\textbf{3.272}^{\pm.021}$ & ${10.337}^{\pm.122}$ & $15.843^{\pm.742}$ & $7.156^{\pm.048}$ & $\textbf{1.351}^{\pm.003}$\\ \hline
    \end{tabular}
}
\label{table:longterm_kitml}
\end{table}

\noindent\textbf{Results.}
In our comparative analysis, detailed in \Cref{table:longterm_hml3d,table:longterm_kitml}, we established the number of actions for generation at $N\!=\!4$. Our model demonstrates superior performance in terms of FID, R-Top3, and MMdist. It exhibits a Jerk value smaller than that of concatenated real motions and approaches the value of individual real motions. These results suggest that our model is capable of generating multi-motion with high fidelity while maintaining continuity and consistency throughout the entire motions.
In \Cref{table:longterm_hml3d}, PriorMDM falls short in FID and R-precision metrics. It also tends to oversmoothens at the transition, resulting in Jerk metric lower than both real single-motion and concatenated real motions. This indicates that the motions generated by PriorMDM lack the depth and subtle characteristics of real motions.
A qualitative comparison in \cref{fig:qualitative} further demonstrates our model's superiority, showcasing more natural, continuous motion compared to T2M-GPT. This quantitative and qualitative evidence underscores our model's advanced capability in producing realistic, coherent long-term human motions.



\begin{table}[t]
\centering
\caption{Single-motion generation performance on HumanML3D. The figures highlighted in \textbf{bold} and \textbf{\bl{blue}} denote the best and second-best results, respectively.}
\resizebox{.95\linewidth}{!}{
    \addtolength{\tabcolsep}{8pt}
    \begin{tabular}{l c c c c c}
        \hline
        Methods & R-Top 3$\uparrow$ & FID$\downarrow$ & MM-Dist$\downarrow$ & Diversity$\rightarrow$ & MModality$\uparrow$ \\
        \hline\hline
        Ground Truth                                         & $0.797^{\pm.002}$ & $0.002^{\pm.000}$ & $2.974^{\pm.008}$ & $9.503^{\pm.065}$ & - \\
        VQ-VAE (reconstruction) & $0.785^{\pm.002}$ & $0.070^{\pm.001}$ & $3.072^{\pm.009}$ & $9.593^{\pm.079}$ & - \\
        \hline
        TEMOS \cite{petrovich2022temos}                     & $0.722^{\pm.002}$ & $3.734^{\pm.028}$ & $3.703^{\pm.008}$ & $0.725^{\pm.071}$ & $0.368^{\pm.018}$\\
        TM2T \cite{guo2022tm2t}                             & $0.729^{\pm.002}$ & $1.501^{\pm.017}$ & $3.467^{\pm.011}$ & $8.973^{\pm.076}$ & $2.424^{\pm.093}$\\
        MotionDiffuse \cite{zhang2022motiondiffuse}         & $0.782^{\pm.001}$ & $0.630^{\pm.001}$ & $3.113^{\pm.001}$ & $\textbf{9.410}^{\pm.049}$ & $1.553^{\pm.042}$\\
        MLD \cite{chen2023executing}                        & $0.736^{\pm.002}$ & $1.087^{\pm.021}$ & $3.347^{\pm.008}$ & $8.589^{\pm.083}$ & $2.219^{\pm.074}$\\
        T2M-GPT \cite{zhang2023generating}                  & $0.775^{\pm.002}$ & $0.116^{\pm.004}$ & $3.118^{\pm.011}$ & $9.761^{\pm.081}$ & $1.856^{\pm.011}$\\
        AttT2M \cite{zhong2023attt2m} & $0.786^{\pm.006}$ & $0.112^{\pm.006}$  & $3.038^{\pm.007}$ & $9.700^{\pm.090}$ & $\textbf{\bl{2.452}}^{\pm.051}$\\
        M2DM \cite{kong2023priority}                        & $0.763^{\pm.003}$ & $0.352^{\pm.005}$ & $3.134^{\pm.010}$ & $9.926^{\pm.073}$ & $\textbf{3.587}^{\pm.072}$\\ \hline
        \textbf{M2D2M (w/ $\beta_t$)}                                & $\textbf{\bl{0.796}}^{\pm.002}$ & $\textbf{\bl{0.115}}^{\pm.006}$ & $\textbf{\bl{3.036}}^{\pm.008}$ & $9.680^{\pm.074}$ & $2.193^{\pm.077}$\\
        \textbf{M2D2M (w/ $\beta(t,d)$)}                                & $\textbf{0.799}^{\pm.002}$ & $\textbf{0.087}^{\pm.004}$ & $\textbf{3.018}^{\pm.008}$ & $\textbf{\bl{9.672}}^{\pm.086}$ & $2.115^{\pm.079}$\\
        \hline  
    \end{tabular}
}
\label{table:single_hml3d}
\end{table}

\begin{table}[t]
\centering
\caption{Single-motion generation performance on KIT-ML. The figures highlighted in \textbf{bold} and \textbf{\bl{blue}} denote the best and second-best results, respectively.}
\resizebox{.95\linewidth}{!}{
    \addtolength{\tabcolsep}{8pt}
    \begin{tabular}{l c c c c c}
        \hline
        Methods & R-Top3$\uparrow$ & FID$\downarrow$ & MM-Dist$\downarrow$ & Diversity$\rightarrow$ & MModality$\uparrow$ \\
        \hline\hline
        Ground Truth                                         & $0.779^{\pm.006}$ & $0.031^{\pm.004}$ & $2.788^{\pm.012}$ & $11.08^{\pm.097}$ & - \\
        VQ-VAE (reconstruction) & $0.740^{\pm.006}$ & $0.472^{\pm.011}$ & $2.986^{\pm.027}$ & $10.994^{\pm.120}$ & - \\
        \hline

        TEMOS \cite{petrovich2022temos}                     & $0.687^{\pm.002}$ & $3.717^{\pm.028}$ & $3.417^{\pm.008}$ & $10.84^{\pm.004}$ & $0.532^{\pm.018}$  \\

        TM2T \cite{guo2022tm2t}                             & $0.587^{\pm.005}$ & $3.599^{\pm.051}$ & $4.591^{\pm.019}$ & $9.473^{\pm.100}$ & $\textbf{\bl{3.292}}^{\pm.034}$  \\
        
        MotionDiffuse \cite{zhang2022motiondiffuse}         & $0.739^{\pm.004}$ & $1.954^{\pm.062}$ & $\textbf{2.958}^{\pm.005}$ & $\textbf{11.10}^{\pm.143}$ & $0.730^{\pm.013}$  \\
        MLD \cite{chen2023executing}                        & $0.734^{\pm.007}$ & $\textbf{\bl{0.404}}^{\pm.027}$ & $3.204^{\pm.027}$ & $10.80^{\pm.117}$ & $2.192^{\pm.071}$  \\
        T2M-GPT \cite{zhang2023generating}                  & $0.737^{\pm.006}$ & $0.717^{\pm.041}$ & $3.053^{\pm.026}$ & $\textbf{\bl{10.862}}^{\pm.094}$ & $1.912^{\pm.036}$ \\
        AttT2M \cite{zhong2023attt2m} & $\textbf{\bl{0.751}}^{\pm.006}$ & $0.870^{\pm.039}$ & $3.309^{\pm.021}$ & $10.96^{\pm.123}$ & $2.281^{\pm.047}$ \\
        M2DM \cite{kong2023priority}                         & ${0.743}^{\pm.004}$ & $0.515^{\pm.029}$ & $3.015^{\pm.017}$ & $11.417^{\pm.097}$ & $\textbf{3.325}^{\pm037}$ \\\hline
        \textbf{M2D2M (w/ $\beta_t$)}                                & ${0.743}^{\pm.006}$ & $\textbf{\bl{0.404}}^{\pm.022}$ & $3.018^{\pm.019}$ & ${10.749}^{\pm.102}$ & $2.063^{\pm.066}$ \\
        \textbf{M2D2M (w/ $\beta(t,d)$)}                                & $\textbf{0.753}^{\pm.006}$ & $\textbf{0.378}^{\pm.023}$ & $\textbf{\bl{3.012}}^{\pm.021}$ & ${10.709}^{\pm.121}$ & $2.061^{\pm.067}$ \\
        \hline
    \end{tabular}
}
\label{table:single_kitml}
\end{table}

\subsection{Single-Motion Generation} \label{sec:single_motion_exp} 
\noindent\textbf{Results.} The task of single-motion generation involves generating human motion from individual action descriptions. In \Cref{table:single_hml3d,table:single_kitml}, we present a comparative analysis of single-motion generation performance against selected baselines. We present comprehensive comparison in Appendix \rd{D}. Our approach outperforms current state-of-the-art methods on both the HumanML3D and KIT-ML datasets, particularly excelling in FID and R-Top 3 metrics. While not leading but closely competitive in other metrics such as MM-Distance, Diversity our method demonstrates robust near-best performance. In terms of multi-modality, our model exhibits lower performance compared to the top-performing models. However, there appears to be a trade-off between multi-modality and FID, suggesting that models with higher FID scores may achieve better multi-modality, as observed in the case of M2DM \cite{kong2023priority} and TM2T \cite{guo2022tm2t}.

\subsection{Ablation Studies} \label{sec:ablation} 
We conduct ablation studies to assess the effects of Dynamic Transition Probability and Two-Phase Sampling on our model. Due to space constraints, we have included further ablation studies in Appendix \rd{C}.

\noindent\textbf{Dynamic Transition Probability.} We first investigate the impact of dynamic transition probability presented in wcref{sec:transition}. In \Cref{table:ablation_multi_hml3d}, we compare the performance of our model with a dynamic token transition probability, denoted as $\beta(t, d)$, as opposed to a static one, $\beta_t$. Dynamic transition probability substantially enhances our model's performance, particularly in terms of FID. In addition, the table shows that it enhances the smoothness proven by the lowest jerk when it is combined with two-phase sampling. This improvement underscores the importance of the synergistic effect of TPS and $\beta(t, d)$.

\noindent\textbf{Two-Phase Sampling (TPS).} To compare multi-motion generation algorithms illustrated in \cref{fig:algorithm}, we evaluate the performance of these algorithms on VQ-Diffusion model in \Cref{table:ablation_tps}, highlighting the effectiveness of TPS in multi-motion generation. TPS significantly improves the smoothness of long-term motion sequences while maintaining fidelity to individual motions, as evidenced by the improved Jerk and FID metrics. However, the results for Handshake and SLERP show an over-smoothing effect when compared to the original dataset, with their Jerk values being lower than that of the single ground truth. Notably, SLERP even exhibits a negative Jerk value, indicating excessive smoothing.

\begin{table}[t]
\centering
\caption{Ablation studies on multi-motion generation performance on HumanML3D. `Individual Motion' denotes individual motions within our motion boundaries.}
\resizebox{0.95\linewidth}{!}{
    \addtolength{\tabcolsep}{8pt}
    \begin{tabular}{c|cccc|ccc}
        \toprule
        \multirow{2}{*}{Methods} & \multicolumn{4}{c}{Individual Motion} & \multicolumn{3}{|c}{Transition (40 frames)}\\\cline{2-5}\cline{5-8}
         & R-Top3$\uparrow$ & FID$\downarrow$ & MMdist$\downarrow$ & Div$\rightarrow$ & FID$\downarrow$ & Div$\rightarrow$ & Jerk$\rightarrow$ \\
        \hline\hline
         Ground Truth (Single) & $0.791$ & $0.002$ & $2.707$ & $9.820$ & $0.003$  & $9.574$ & $1.192$    \\
         Ground Truth (Concat) & - & - & - & - & - & - & $1.371$    \\
         \midrule
         $\beta_t, T_s=100$ & $0.749$ & $0.212$ & $3.015$  & $9.990$ & $3.324$ & $8.681$  & $1.248$  \\
         $\beta_t, T_s=90$ & $0.738$ & $0.253$ & $3.164$  & $9.822$ & $3.483$ & $\textbf{8.625}$  & $1.265$  \\
         $\beta(t,d), T_s=100$ & $0.751$ & $0.196$ & $3.012$ & $9.894$  & $3.340$ & $8.751$ & $1.248$  \\
         $\beta(t,d), T_s=90$ & $0.733$ & $0.253$ & $ 3.165$ & $9.806$ & $\textbf{3.276}$  & $8.599$  & $\textbf{1.238}$ \\
         \bottomrule
    \end{tabular}
    }
\label{table:ablation_multi_hml3d}
\end{table}

\begin{table}[!t]
\centering
\caption{Multi-motion generation on different smoothing methods on HumanML3D.}
    \addtolength{\tabcolsep}{8pt}
    \resizebox{0.95\linewidth}{!}{
    \begin{tabular}{c | c c c c|c c c}
        \toprule
         \multirow{2}{*}{Methods} & \multicolumn{4}{c}{Individual Motion} & \multicolumn{3}{|c}{Transition (40 frames)}\\\cline{2-5}\cline{6-8}
         & R-Top3$\uparrow$ & FID$\downarrow$ & MMdist$\downarrow$ & Div$\rightarrow$ & FID$\downarrow$ & Div$\rightarrow$ & Jerk$\rightarrow$ \\
        \hline\hline
        Ground Truth (Single) & $0.791$ & $0.002$ & $2.707$ & $9.820$ & $0.003$ & $9.574$ & $1.192$ \\
        Ground Truth (Concat) & - & - & - & - & - & - & $1.371$ \\
        \midrule
        Handshake \cite{shafir2023human} & $0.635$ & $1.279$ & $4.182$ & $8.939$ & $\textbf{3.039}$ & $8.566$ & $1.097$ \\
        SLERP \cite{TEACH:3DV:2022} & 0.549 & 1.402 & 4.679 & 8.535 & $4.873$ & $7.912$ & $-3.554$ \\\textbf{TPS (Ours)} & $\textbf{0.733}$ & $\textbf{0.254}$ & $ \textbf{3.165} $ & $\textbf{9.806}$ & $\underline{3.276}$ & $\textbf{8.599}$ & $\textbf{1.238}$ \\
 
        \bottomrule  
    \end{tabular}
    }
    \label{table:ablation_tps}
\end{table}



%% file: sec/6_conclusion.tex
\section{Conclusion \& Discussion} \label{sec:conclusion}
We present M2D2M, a model designed for generating multi-motion sequences from a set of action descriptions. Incorporating a Dynamic Transition Matrix and Two-Phase Sampling, M2D2M achieves state-of-the-art performance in generating human motion from text tasks. For multi-motion generation, a ground truth is absent as we create extended sequences from multiple action descriptions. Consequently, we introduce a new evaluation metric to assess the smoothness of the motion. However, these metrics do not comprehensively evaluate the generated multi-motion sequences, as they do not account for all possible scenarios. Addressing this limitation remains for future work. Additionally, our research aims to enhance virtual reality and assistive technologies but could raise privacy and security concerns, requiring strict data policies and transparent monitoring.


%% file: sec/X_suppl.tex
\clearpage
\setcounter{page}{1}
\appendix

\title{Supplementary Material of M2D2M: Multi-Motion Generation from Text with Discrete Diffusion Models}

\titlerunning{Supplementary Material of M2D2M}

\author{Seunggeun Chi\inst{1,2}$^{*}$ \and
Hyung-gun Chi\inst{2}$^{*}$\and
Hengbo Ma$^{\ddag}$\and
Nakul Agarwal\inst{1}\and \\
Faizan Siddiqui\inst{1}\and
Karthik Ramani\inst{2}$^\dag$\and
Kwonjoon Lee\inst{1}$^\dag$
}

\authorrunning{S.~Chi et al.}

\institute{\hspace{-0.5em}Honda Research Institute USA \and \hspace{-0.5em}Purdue University \\
\email{\{chi65, chi45\}@purdue.edu},\\
\email{ramani@purdue.edu}, \email{kwonjoon\_lee@honda-ri.com}}

\footnotetext[1]{Co-first authors.} 
\footnotetext[2]{Senior authors.}
\footnotetext[3]{Work done at Honda Research Institute.}

\maketitle

In the supplementary material, we offer additional details and experiments that are not included in the main paper due to the page limit. This includes implementation specifics and architectural design, along with baseline implementation methodologies (\cref{sec:imple_detail}). Additionally, we describe the generation of test sets for the multi-motion generation task in \cref{sec:testset}, present further ablation studies in \cref{sec:additional_ablations}, report comprehensive comparison in \cref{sec:detailed}, and provide in-depth analysis of our work in \cref{sec:analysis}. Lastly, we include extra qualitative results in \cref{sec:addtional_qualitative}.

\section{Additional Details} \label{sec:imple_detail}
\subsection{M2D2M}
\noindent\textbf{Motion VQ-VAE.} In developing the Motion VQ-VAE, we adopt the architecture proposed by Zhang \etal \cite{zhang2023generating}. We construct both the encoder and decoder of the Motion VQ-VAE using a CNN-based architecture, specifically employing 1D convolutions. Additionally, we adhere to the same hyperparameters and training procedures as outlined in their study.

\begin{figure}[t!]
\centering
\includegraphics[width=0.6\linewidth]{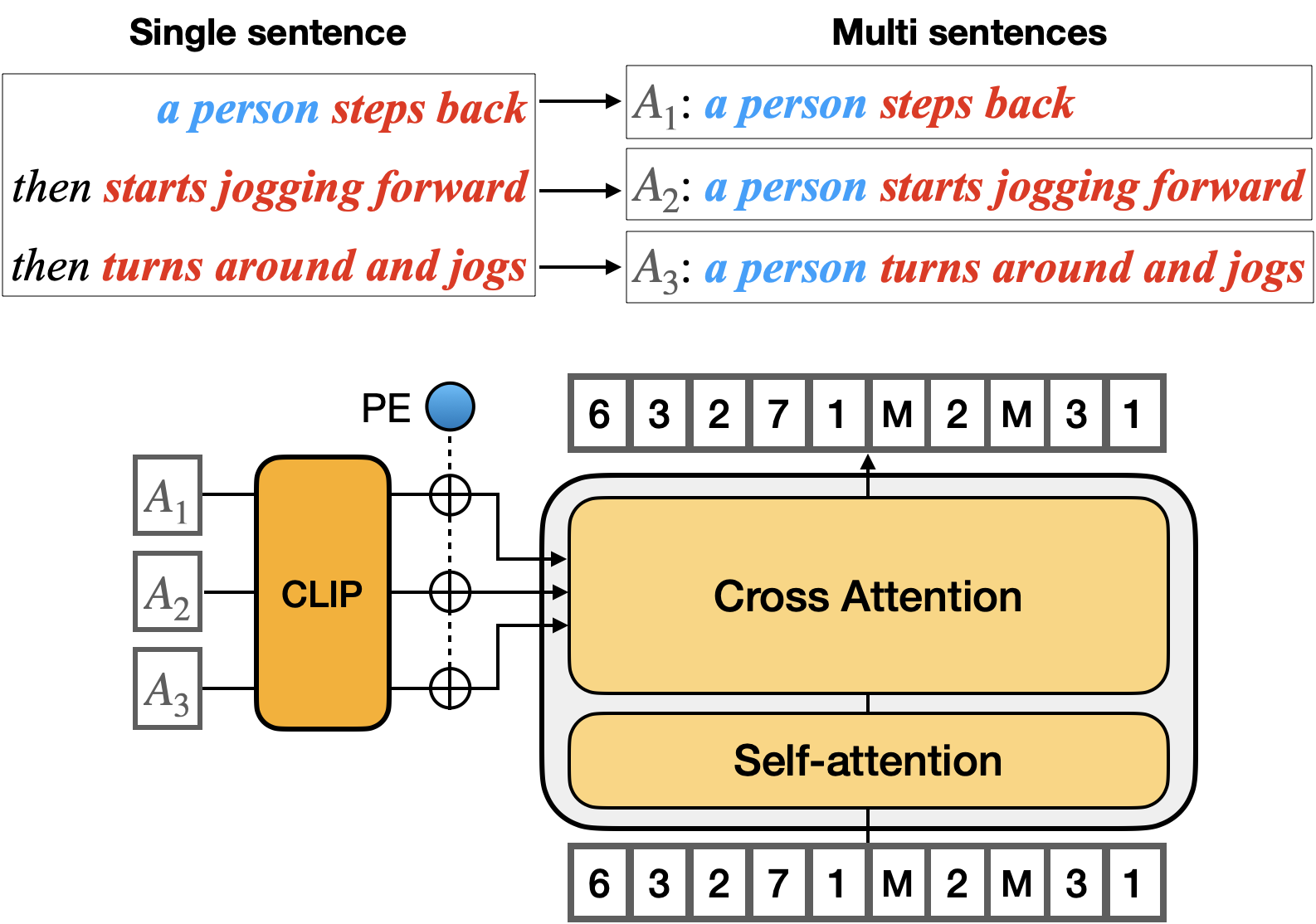}
\caption{Overview of action sentence conditioning of M2D2M. We initially decompose sentences to extract action verbs and subsequently utilize these verbs to construct new sentences. These newly formed sentences then serve as conditions for generating human motion sequences.}
\label{fig:action_cond}
\vspace{-1em}
\end{figure}

\vspace{.3em}
\noindent\textbf{Denoising Transformer.} The denoising transformer configuration is specified as follows: 12 layers, 16 attention heads, 512 embedding dimensions, 2048 hidden dimensions, and a dropout rate of 0.
Also, we designed action sentence conditioning for the denoising transformer to enable the multi-motion generation task with the HumanML3D dataset and KIT-ML dataset. We focus on the action verbs within a sentence (i.e., `walk', `turn around') of datasets, because they offer clear information about the type of motion involved. Therefore, we further break down the sentence using action verbs and then enrich them to form a complete action description, like `a person walking,' which serves as the basis for conditioning the motion generation as illustrated in \cref{fig:action_cond}. For a joint sampling of Two-Phase Sampling (TPS), which aims to create a seamless motion sequence, we concatenate action tokens from successive actions for conditioning. This forms a compound condition that infuses the motion generation with contextual information, ensuring the resulting sequence is both cohesive and reflective of the intended actions.

\vspace{.3em}
\noindent\textbf{Implementation Details.}
Our model adheres to the hyper-parameter settings of VQ-Diffusion \cite{gu2022vector} unless otherwise stated, encompassing the configurations for the transition matrix parameters, namely $\bar{\alpha}_t$ and $\bar{\gamma}_t$. We linearly increase the $\bar{\gamma}_t$ and decrease the $\bar{\alpha}_t$.
The loss coefficient is set at $\lambda = 5.0 \times 10^{-4} $ as per \cref{eq:vq_diffusion}, and the diffusion process is defined over $T = 100$ timesteps. Optimization is carried out using the AdamW optimizer with a learning rate of $2.0 \times 10^{-4}$, $\beta_1 = 0.9$, $\beta_2 = 0.99$, and weight decay $4.5 \times 10^{-2}$. We trained the model for 110 epochs, and the learning rate decayed to $2.0 \times 10^{-5}$ at the 100th epoch. We use the guidance scale of $s=4$ for single motion generation, and $s=2$ for multi-motion generation. When generating multi-motion sequence, we use $T_s=90$ for TPS. For generating single motions, we apply a Dynamic Transition Probability scale factor of $\eta = 0.5$, and for multi-action generation, we adjust the scale factor to $\eta = 0.25$.

\subsection{Baselines for Multi-Motion Generation}
We evaluated the baseline methods of T2M-GPT\footnote{\url{https://github.com/Mael-zys/T2M-GPT}} \cite{zhang2023generating} and PriorMDM\footnote{\url{https://githubwcom/priorMDM/priorMDM}} \cite{shafir2023human} for the task of multi-motion generation based on the code provided from the original papers. For a fair comparison with T2M-GPT, we modified the model to produce codebooks matching the specified ground truth length by disabling the end-token output. These codebooks were then concatenated for each motion and fed into the decoder.
In the case of PriorMDM and Handshake \cite{shafir2023human}, we set the hyper-parameter to match the illustration of \cref{fig:algorithm} in the main paper for the fair comparison, employing a handshake size of 40 and transition margins of 20. For the other hyper-parameters, we follow the setup of PriorMDM \cite{shafir2023human}.
For the SLERP algorithm, unlike the TEACH \cite{TEACH:3DV:2022} setup, we first independently generate individual motions with half-transition length shorter than the given ground truth length, then apply SLERP as illustrated in \cref{fig:algorithm} of the main paper.
We computed the FID score based on their prescribed method, for both individual motions and transitions.

\section{Multi-Motion Generation Test Set} \label{sec:testset}
Due to the absence of distinct motion boundaries in multi-action verb annotations within the HumanML3D and KIT-ML datasets used in our experiments, we opted for test sets that exclusively consist of single action verbs. In the curated test sets, each sentence includes only one action verb, such as `walk' or `run'. We then randomly selected $N$ action descriptions from this pool of single-action verb sentences, ensuring no overlap, to create our test set for the multi-motion generation task. Specifically, for $N=4$, the test set from the HumanML3D dataset comprises 1448 motions, each associated with a single-verb annotation. Similarly, the test set from the KIT-ML dataset includes 532 motions, all characterized by single action verb annotations.

\section{Additional Ablation Studies} \label{sec:additional_ablations}
In this section, we present a series of additional ablation studies that were not included in \cref{sec:ablation} of the main paper due to the page limit. It includes 1) exploring different classifier-free guidance scales (\cref{sec:ablation_s}), 2) assessing our model's performance with varying numbers of actions in multi-motion generation tasks (\cref{sec:ablation_N}), 3) examining the smoothness-fidelity trade-off at different independent sampling steps in TPS (\cref{sec:ablation_Ts}), and finally, 4) evaluating the Dynamic Transition Probability scale $\eta$ (\cref{sec:TPM}).

\begin{table}[!t]
\centering
\caption{Comparison table for Multi-motion generation performance with different classifier-free scales on HumanML3D dataset.}

\resizebox{1.0\linewidth}{!}{
    \addtolength{\tabcolsep}{4pt}
    \begin{tabular}{c|cccc|ccc}
        \hline
        \multirow{2}{*}{\begin{tabular}[c]{@{}c@{}}Classifier-free\\ Guidance Scale ($s$)\end{tabular}} & \multicolumn{4}{c}{Individual Motion} & \multicolumn{3}{c}{Transition (40 frames)} \\
        \cline{2-5} \cline{6-8}
         & R-Top3$\uparrow$ & FID$\downarrow$ & MMdist$\downarrow$ & Div$\rightarrow$ & FID$\downarrow$ & Div$\rightarrow$ & Jerk$\rightarrow$ \\
        \hline\hline
        Ground Truth (Single) & $0.791^{\pm.002}$ & $0.002^{\pm.000}$ & $2.707^{\pm.008}$ & $9.820^{\pm.065}$ & $0.003^{\pm.002}$ & $9.574^{\pm.054}$ & $1.192^{\pm.005}$ \\
        Ground Truth (Concat) & - & - & - & - & - & - & $1.371^{\pm{.004}}$ \\ \hline
        1.0 & $0.628^{\pm.005}$ & $0.350^{\pm.021}$ & $3.836^{\pm.019}$ & $9.573^{\pm.156}$ & $3.299^{\pm.152}$ & $8.395^{\pm.142}$ & $1.246^{\pm.006}$ \\
        1.5 & $0.705^{\pm.004}$ & $\textbf{0.254}^{\pm.017}$ & $3.063^{\pm.017}$ & $9.777^{\pm.170}$ & $3.293^{\pm.177}$ & $8.545^{\pm.115}$ & $1.242^{\pm.009}$ \\
        2.0 & $0.733^{\pm.003}$ & $\textbf{0.254}^{\pm.016}$ & $3.165^{\pm.019} $ & $\textbf{9.806}^{\pm.158}$ & $\textbf{3.276}^{\pm.173}$ & $8.599^{\pm.154}$ & $\textbf{1.238}^{\pm.008}$ \\
        2.5 & $0.746^{\pm.006}$ & $0.262^{\pm.025}$ & $3.063^{\pm.017}$  & ${9.844}^{\pm.148}$ & $3.321^{\pm.178}$ & $8.622^{\pm.124}$ & $1.252^{\pm.009}$  \\
        3.0 & $\textbf{0.751}^{\pm.006}$ & $0.270^{\pm.020}$ & $\textbf{3.042}^{\pm.023}$ & ${9.795}^{\pm.147}$ & ${3.400}^{\pm.194}$ & $\textbf{8.648}^{\pm.130}$ & $1.263^{\pm.007}$  \\\hline
    \end{tabular}
}
\vspace{-1em}
\label{table:classifier_free}
\end{table}

\begin{table}[t]
\centering
\caption{Single-motion generation performance on the different classifier-free guidance scale on HumanML3D.}
\resizebox{0.7\linewidth}{!}{
    \addtolength{\tabcolsep}{+4pt}
    \begin{tabular}{c | c c c c}
        \hline
        \multicolumn{1}{c|}{\begin{tabular}[c]{@{}c@{}}Classifier-free\\ Guidance Scale ($s$)\end{tabular}} & R-Top 3$\uparrow$ & FID$\downarrow$ & MM-Dist$\downarrow$ & Diversity$\rightarrow$ \\
        \hline\hline
        Ground Truth                                         & $0.797^{\pm.002}$ & $0.002^{\pm.000}$ & $2.974^{\pm.008}$ & $9.503^{\pm.065}$\\
        \hline
        \textbf{$0.0$}                                & $0.686^{\pm.003}$ & $0.107^{\pm.005}$ & $3.690^{\pm.008}$ & $9.580^{\pm.088}$\\
        \textbf{$1.0$}                                & $0.786^{\pm.003}$ & $0.146^{\pm.002}$ & $3.084^{\pm.008}$ & $9.897^{\pm.088}$\\
        \textbf{$2.0$}                                & $\textbf{0.804}^{\pm.003}$ & $0.139^{\pm.004}$ & $2.995^{\pm.008}$ & $9.886^{\pm.082}$\\
        \textbf{$3.0$}                                & $0.803^{\pm.002}$ & $0.107^{\pm.003}$ & $\textbf{2.980}^{\pm.006}$ & $9.815^{\pm.089}$\\
        \textbf{$4.0$}                                & $0.799^{\pm.002}$ & $\textbf{0.087}^{\pm.003}$ & $3.018^{\pm.008}$ & $9.672^{\pm.086}$\\
        \textbf{$5.0$}                                & $0.787^{\pm.002}$ & $0.127^{\pm.007}$ & $3.089^{\pm.007}$ & 
        $\textbf{9.439}^{\pm.086}$\\
        \hline
    \end{tabular}
    \addtolength{\tabcolsep}{-2pt}
}
\label{table:classifier_free}
\vspace{-1em}
\end{table}

\begin{table}[!t]
\centering
\caption{Single motion generation performance on different distance functions for $d(\cdot,\cdot)$ on Human3D dataset.}
\vspace{-1em}
\resizebox{1.0\linewidth}{!}{
    \addtolength{\tabcolsep}{7pt}
    \begin{tabular}{lccccc}
        \hline
        Methods &  R-Top3$\uparrow$  & FID$\downarrow$ & MM-Dist$\downarrow$ & Diversity$\rightarrow$ & MModality$\uparrow$\\
        \hline\hline
        L2 & $0.798^{\pm.002}$ & $0.098^{\pm.005}$ & $3.018^{\pm.008}$ & $\textbf{9.623}^{\pm.085}$ & $2.115^{\pm.079}$\\
        L2 Rank& $0.799^{\pm.002}$ & $\textbf{0.087}^{\pm.004}$ & $3.018^{\pm.008}$ & $9.672^{\pm.086}$ & $2.132^{\pm.073}$\\ \hline
        Cosine & $\textbf{0.801}^{\pm.002}$ & $0.092^{\pm.004}$ & $\textbf{3.011}^{\pm.008}$ & $9.670^{\pm.084}$ & $\textbf{2.137}^{\pm.084}$\\ 
        Cosine Rank & $0.797^{\pm.002}$ & $0.099^{\pm.005}$ & $3.026^{\pm.008}$ & ${9.669}^{\pm.085}$ & $2.125^{\pm.069}$\\
        \hline
    \end{tabular}
}
\vspace{-1em}
\label{table:distance}
\end{table}


\subsection{Classifier-free Guidance Scale} \label{sec:ablation_s}
We first focus on the effect of different classifier guidance scales $s$, which is described in \cref{eq:classifier_free}. To evaluate the performance of our model in multi-motion generation and single-motion generation, we utilize the HumanML3D dataset, and provide results presented in \Cref{table:classifier_free}.  This experiment reveals that the optimal balance between accuracy and fidelity for these metrics is achieved at a classifier guidance scale of $s=4$ for single-motion generation, and best smoothness at $s=2$ for multi-motion generation.

\subsection{Number of Action in Multi-Motion Generation} \label{sec:ablation_N}
In order to explore our model's effectiveness in generating long-term motion, we evaluate the performance of our model by progressively increasing the number of actions ($N$) using the HumanML3D dataset. The results of these evaluations are detailed in \Cref{table:num_act}. We found that as $N$ increases, R-Top3 and FID scores of individual motion demonstrate a decline, indicating a reduction in fidelity with more actions. Despite this, it's noteworthy that our model's performance on the transition part remains comparably effective to that of real single motions, even at $N=32$, a considerably long motion sequence. This highlights our model's proficiency in generating long-term motion with smooth and coherent transitions.

\subsection{Different Distance metrics for Dynamic Transition Probability} In \Cref{table:distance}, we conduct a comparative analysis of different distance functions for $d(\cdot, \cdot)$, utilized in defining the codebook distance for \cref{eq:q_t}. Specifically, we evaluate the performance of L2 and Cosine Distance, focusing on their effectiveness as distance functions. Our findings indicate that the L2 Rank distance function yields the best FID score, highlighting its superiority in this context.



\begin{table}[t]
\centering
\caption{Multi-motion generation performance on the different number of actions ($N$) on HumanML3D.}
\resizebox{1.0\linewidth}{!}{
    \addtolength{\tabcolsep}{4pt}
    \begin{tabular}{c|cccc|ccc}
        \hline
        \multirow{2}{*}{{\begin{tabular}[c]{@{}c@{}}The number\\of actions $(N)$ \end{tabular}}} & \multicolumn{4}{c}{Individual Motion} & \multicolumn{3}{c}{Transition (40 frames)}\\\cline{2-5}\cline{6-8}
         & R-Top3$\uparrow$ & FID$\downarrow$ & MMdist$\downarrow$ & Div$\rightarrow$ & FID$\downarrow$ & Div$\rightarrow$ & Jerk$\rightarrow$  \\
        \hline\hline
        Ground Truth (Single) & $0.791^{\pm.002}$ & $0.002^{\pm.000}$ & $2.707^{\pm.008}$ & $9.820^{\pm.065}$ & $0.003^{\pm.002}$ & $9.574^{\pm.054}$ & $1.192^{\pm.005}$ \\
        Ground Truth (Concat) & - & - & - & - & - & - & $1.371^{\pm{.004}}$ \\ \hline
        $N=1$ & $0.751^{\pm.008}$ & $0.196^{\pm.003}$ & ${3.012}^{\pm.018}$ & $9.894^{\pm.057}$  & $3.340^{\pm.219}$ & $8.751^{\pm.005}$ & $1.248^{\pm.005}$ \\
        $N=2 $  & ${0.737}^{\pm.007}$ & ${0.198}^{\pm.025}$ & ${3.127}^{\pm.031} $ & ${9.870}^{\pm.064}$ & ${3.430}^{\pm.431}$ & $8.497^{\pm.121}$ & ${1.244}^{\pm.013}$ \\
        $N=4 $  & ${0.733}^{\pm.003}$ & ${0.254}^{\pm.016}$ & ${3.165}^{\pm.019} $ & ${9.806}^{\pm.158}$ & ${3.276}^{\pm.173}$ & $8.599^{\pm.154}$ & ${1.238}^{\pm.008}$ \\
        $N=8 $  & $0.733^{\pm.005}$ & $0.307^{\pm.027}$ & $3.153^{\pm.028}$ & $9.624^{\pm.137}$ & $3.343^{\pm.092}$ & $8.675^{\pm.121}$ & $1.255^{\pm.010}$\\
        $N=16$  & $0.725^{\pm.004}$ & $0.312^{\pm.031}$ & $3.193^{\pm.018}$ & $9.557^{\pm.066}$ & $3.380^{\pm.109}$ & $8.455^{\pm.165}$ & $1.245^{\pm.011}$\\
        $N=32$  & $0.731^{\pm.005}$ & $0.350^{\pm.040}$ & $3.192^{\pm.023}$ & $9.555^{\pm.069}$ & $3.336^{\pm.145}$ & $8.537^{\pm.182}$ & $1.248^{\pm.013}$\\
        \hline
    \end{tabular}
    }
\label{table:num_act}
\end{table}

\begin{table}[t]
\centering
\caption{Multi-motion generation performance across a different number of independent denoising steps ($T_s$) of Two-Phase Sampling on HumanML3D.}
\resizebox{1.0\linewidth}{!}{
    \addtolength{\tabcolsep}{4pt}
    \begin{tabular}{c|cccc|ccc}
        \hline
        \multirow{2}{*}{Methods} & \multicolumn{4}{c|}{Individual Motion} & \multicolumn{3}{c}{Transition (40 frames)} \\\cline{2-5}\cline{6-8}
         & R-Top3$\uparrow$ & FID$\downarrow$ & MMdist$\downarrow$ & Div$\rightarrow$ & FID$\downarrow$ & Div$\rightarrow$ & Jerk$\rightarrow$ \\
        \hline\hline
        Ground Truth (Single) & $0.791^{\pm.002}$ & $0.002^{\pm.000}$ & $2.707^{\pm.008}$ & $9.820^{\pm.065}$ & $0.003^{\pm.002}$ & $9.574^{\pm.054}$ & $1.192^{\pm.005}$ \\
        Ground Truth (Concat) & - & - & - & - & - & - & $1.371^{\pm{.004}}$ \\ \hline
          w/o TPS   & $\textbf{0.755}^{\pm.007}$ & $\textbf{0.173}^{\pm.010}$ & $3.015^{\pm.024}$ & $9.950^{\pm.076}$ & $3.455^{\pm.142}$ & $8.554^{\pm.081}$ & $1.402^{\pm.005}$ \\
          $T_s=100$ & $0.751^{\pm.008}$ & $0.196^{\pm.003}$ & $\textbf{3.012}^{\pm.018}$ & $9.894^{\pm.057}$  & $3.340^{\pm.219}$ & $8.751^{\pm.005}$ & $1.248^{\pm.005}$ \\
          $T_s=95$  & $0.737^{\pm.004}$ & $0.232^{\pm{.028}}$ & $3.105^{\pm{.017}}$ & $9.772^{\pm.167}$ & $3.289^{\pm.243}$ & $8.643^{\pm{.132}}$ &$1.253^{\pm.007}$\\
          $T_s=90$  & $0.733^{\pm.003}$ & $0.254^{\pm.016}$ & $3.165^{\pm.019} $ & $\textbf{9.806}^{\pm.158}$ & $\textbf{3.276}^{\pm.173}$ & $8.599^{\pm.154}$ & $\textbf{1.238}^{\pm.008}$ \\
          $T_s=80$  & ${0.725}^{\pm.006}$ & ${0.284}^{\pm.024}$ & ${3.194}^{\pm.029}$ & $9.767^{\pm.129}$ & $3.338^{\pm.129}$ & $\textbf{8.691}^{\pm.114}$ & $1.247^{\pm.007}$\\
          $T_s=50$  & ${0.709}^{\pm.006}$ & ${0.371}^{\pm.034}$ & ${3.315}^{\pm.018}$ & $9.665^{\pm.125}$ & $3.282^{\pm.263}$ & $8.595^{\pm.144}$ & $1.254^{\pm.010}$\\\hline
    \end{tabular}
    }
\label{table:tau}
\end{table}

\begin{table}[!t]
\centering
\caption{Multi-motion generation on different smoothing methods with MDM on HumanML3D.}
\vspace{-1em}
    \addtolength{\tabcolsep}{8pt}
    \resizebox{1.0\linewidth}{!}{
    \begin{tabular}{c | c c c c|c c c}
        \toprule
         \multirow{2}{*}{Methods} & \multicolumn{4}{c}{Individual Motion} & \multicolumn{3}{|c}{Transition (40 frames)}\\\cline{2-5}\cline{6-8}
         & R-Top3$\uparrow$ & FID$\downarrow$ & MMdist$\downarrow$ & Div$\rightarrow$ & FID$\downarrow$ & Div$\rightarrow$ & Jerk$\rightarrow$ \\
        \hline\hline
        Ground Truth (Single) & $0.791$ & $0.002$ & $2.707$ & $9.820$ & $0.003$ & $9.574$ & $1.192$ \\
        Ground Truth (Concat) & - & - & - & - & - & - & $1.371$ \\
        \midrule
        MDM \cite{tevet2022human} + Handshake \cite{shafir2023human} & $0.586$ & $0.832$ & $5.901$ & $\textbf{9.543}$ & $\textbf{3.351}$  & $\textbf{8.801}$ & $0.476$ \\
        MDM \cite{tevet2022human} + \textbf{TPS (Ours)} & $\textbf{0.640}$ & $\textbf{0.582}$ & $\textbf{5.287}$ & $9.321$ & $3.376$ & $8.070$ & $\textbf{0.634}$ \\
 
        \bottomrule  
    \end{tabular}
    }
    \label{table:ablation_tps_regular}
    \vspace{-1.4em}
\end{table}

\subsection{Effect of Two-Phase Sampling} \label{sec:ablation_Ts}
In \Cref{table:tau}, we explore the impact of Two-Phase Sampling. Our analysis also includes adjustments in the ratio of independent denoising steps $(T_s)$ to the total number of denoising steps $(T)$. This examination reveals a clear trade-off in motion generation between smoothness and fidelity. As discussed in \cref{sec:sampling}, phases of independent sampling enhance the fidelity of individual motions, while phases of joint sampling improve the fidelity and smoothness of transitions between motions. Implementing the Two-Phase Sampling algorithm and reducing the number of independent sampling steps $(T_s)$ tends to improve smoothness metrics (e.g., Jerk), but simultaneously, fidelity metrics such as R-Top3 and FID begin to deteriorate. This observation emphasizes the intrinsic trade-off between smoothness and fidelity in motion generation, identifying an optimal $T_s = 90$ for the smoothness metric being identified.

In \Cref{table:ablation_tps_regular}, we evaluate multi-motion generation algorithms on non-latent diffusion models.
We applied Handshake \cite{shafir2023human} and TPS to MDM \cite{tevet2022human}, a diffusion model operating in Cartesian space with 3D skeletal coordinates. 
We observe that the effectiveness of TPS is not confined to its designed latent space; it also functions effectively in the Cartesian domain.
The results show that TPS achieves better FID and R-Precision for individual motions, albeit with reduced diversity.
For the transition part, TPS demonstrates comparable FID results while exhibiting improved smoothness as measured by Jerk.

\subsection{The scale of Transition Probability Matrix} \label{sec:TPM}
We investigated the impact of dynamic transition probability on the generation of multiple motions by conducting an ablation study that varied the transition probability scale, $\eta$. In \Cref{table:eta}, we noted that the dynamic transition probability, $\beta(d,t)$, outperforms the traditional method of $\beta(t)$. Additionally, the results indicate a trend where the smoothness metric (Jerk) becomes closer to ground truth single motion as $\eta$ is reduced.

\begin{table}[t]
\centering
\caption{Multi-motion generation performance across a different number of independent denoising steps ($T_s$) of Two-Phase Sampling on HumanML3D.}
\resizebox{1.0\linewidth}{!}{
    \addtolength{\tabcolsep}{4pt}
    \begin{tabular}{c|cccc|ccc}
        \hline
        \multirow{2}{*}{\begin{tabular}[c]{@{}c@{}}Transition Probability\\ Methods\end{tabular}} & \multicolumn{4}{c}{Individual Motion} & \multicolumn{3}{c}{Transition (40 frames)}\\\cline{2-5}\cline{5-8}
         & R-Top3$\uparrow$ & FID$\downarrow$ & MMdist$\downarrow$ & Div$\rightarrow$ & FID$\downarrow$ & Div$\rightarrow$ & Jerk$\rightarrow$  \\
        \hline\hline
        Ground Truth (Single) & $0.791^{\pm.002}$ & $0.002^{\pm.000}$ & $2.707^{\pm.008}$ & $9.820^{\pm.065}$ & $0.003^{\pm.002}$ & $9.574^{\pm.054}$ & $1.192^{\pm.005}$ \\
        Ground Truth (Concat) & - & - & - & - & - & - & $1.371^{\pm{.004}}$ \\ \hline
         {\begin{tabular}{c|c}$\beta(t)\quad\,\,$&\qquad\ -\quad\,\, \end{tabular}} & $0.738^{\pm.009}$ & $0.253^{\pm.002}$ & $3.164^{\pm.021}$  & $9.822^{\pm.051}$ & $3.483^{\pm.029}$ & $8.625^{\pm.044}$  & $1.265^{\pm.005}$  \\
         {\begin{tabular}{c|c}$\beta(d,t)\quad$&$\quad\eta=1.00$ \end{tabular}} & $0.730^{\pm.005}$ & $0.264^{\pm.026}$ & $3.152^{\pm{.028}}$ & $9.808^{\pm{.162}}$ & $3.315^{\pm{.225}}$ & $8.654^{\pm{0.064}}$ & $1.252^{\pm.007}$\\
         {\begin{tabular}{c|c}$\beta(d,t)\quad$&$\quad\eta=0.50$ \end{tabular}} & $0.733^{\pm.003}$ & $\textbf{0.244}^{\pm.016}$ & $3.156^{\pm.029} $ & $9.830^{\pm.160}$ & $3.278^{\pm.138}$ & $8.586^{\pm.127}$ & $1.250^{\pm.008}$ \\
         {\begin{tabular}{c|c}$\beta(d,t)\quad$&$\quad\eta=0.33$ \end{tabular}}    &$0.732^{\pm004}$ & $0.245^{\pm.010}$ & $\textbf{3.150}^{\pm{.173}}$ & $\textbf{9.815}^{\pm.152}$ & $3.312^{\pm{.171}}$ & $\textbf{8.675}^{\pm{.134}}$ &  $1.246^{\pm.009}$ \\
         {\begin{tabular}{c|c}$\beta(d,t)\quad$&$\quad\eta=0.25$ \end{tabular}}    & $\textbf{0.734}^{\pm.003}$ & $0.253^{\pm.016}$ & $3.165^{\pm.019} $ & $9.806^{\pm.158}$ & $\textbf{3.276}^{\pm.017}$ & $8.599^{\pm.154}$ & $\textbf{1.238}^{\pm.008}$ \\
         {\begin{tabular}{c|c}$\beta(d,t)\quad$&$\quad\eta=0.20$ \end{tabular}}    & $0.724^{\pm005}$ & $0.254^{\pm.010}$ & $3.194^{\pm{.026}}$ & $9.803^{\pm{.152}}$ & $3.330^{\pm{.205}}$ & $8.519^{\pm{.162}}$ &  $1.247^{\pm.008}$  \\\hline
    \end{tabular}
    }
\label{table:eta}
\end{table}

\begin{figure}[!t]
\centering
\includegraphics[width=.45\linewidth]{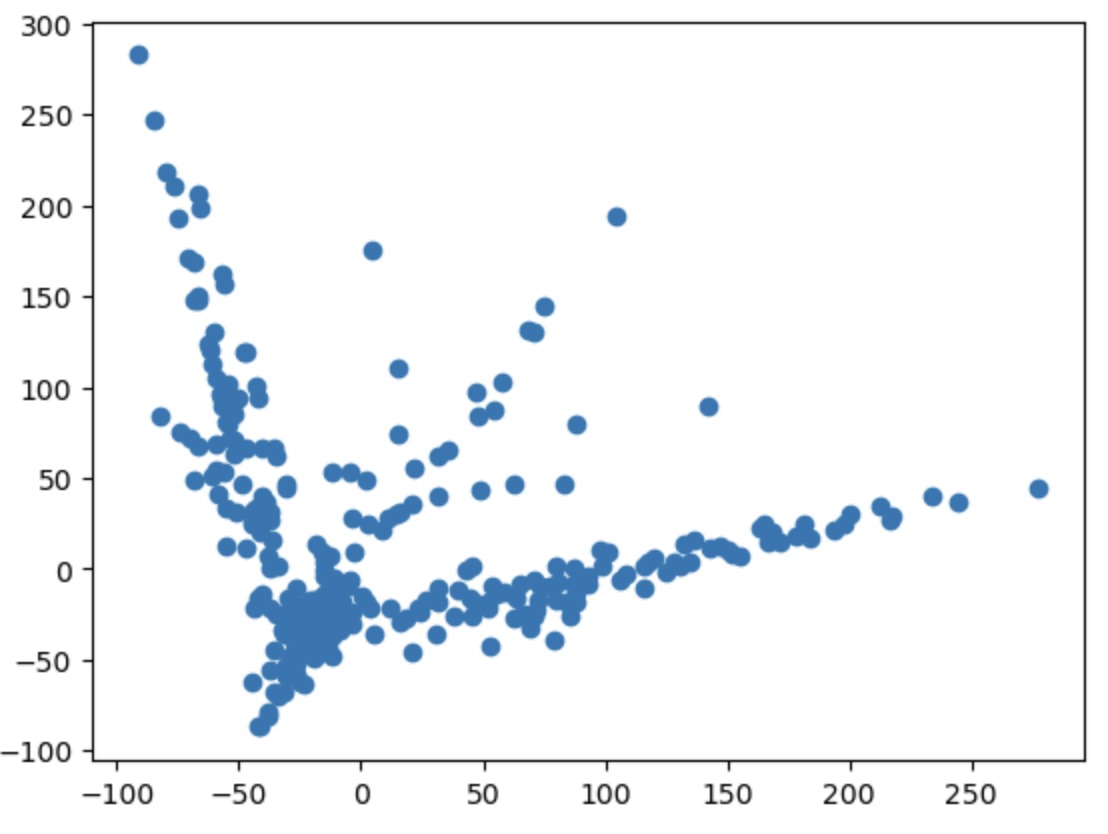}
\includegraphics[width=.45\linewidth]{fig/PCA_3D.png}
\caption{PCA plot representing motion tokens from the codebook of Motion VQ-VAE, visualized in 2D (\textbf{Left}) and 3D (\textbf{Right}) space.}
\label{fig:codebook}
\end{figure}

\section{Comprehensive Comparison of Single-Motion Generation}
\label{sec:detailed}
\begin{table}[!t]
\centering
\caption{Single-motion generation performance on HumanML3D. The figures highlighted in \textbf{bold} and \textbf{\bl{blue}} denote the best and second-best results, respectively.}
\resizebox{.99\linewidth}{!}{
    \addtolength{\tabcolsep}{8pt}
    \begin{tabular}{l c c c c c}
        \hline
        Methods & R-Top 3$\uparrow$ & FID$\downarrow$ & MM-Dist$\downarrow$ & Diversity$\rightarrow$ & MModality$\uparrow$ \\
        \hline\hline
        Ground Truth                                         & $0.797^{\pm.002}$ & $0.002^{\pm.000}$ & $2.974^{\pm.008}$ & $9.503^{\pm.065}$ & - \\
        VQ-VAE (reconstruction) & $0.785^{\pm.002}$ & $0.070^{\pm.001}$ & $3.072^{\pm.009}$ & $9.593^{\pm.079}$ & - \\
        \hline
        Seq2Seq \cite{lin2018generating}                    & $0.396^{\pm.002}$ & $11.75^{\pm.035}$ & $5.529^{\pm.007}$ & $6.223^{\pm.061}$ & -\\
        J2LP \cite{ahuja2019language2pose}                  & $0.486^{\pm.002}$ & $11.02^{\pm.046}$ & $5.296^{\pm.008}$ & $6.223^{\pm.058}$ & -\\
        Text2Gesture \cite{bhattacharya2021text2gestures}   & $0.345^{\pm.002}$ & $5.012^{\pm.030}$ & $6.030^{\pm.008}$ & $7.676^{\pm.071}$ & -\\
        Hier \cite{ghosh2021synthesis}                      & $0.552^{\pm.004}$ & $6.532^{\pm.024}$ & $5.012^{\pm.018}$ & $6.409^{\pm.042}$ & -\\
        MoCoGAN \cite{tulyakov2018mocogan}                  & $0.106^{\pm.001}$ & $94.41^{\pm.021}$ & $9.643^{\pm.006}$ & $8.332^{\pm.008}$ & $0.019^{\pm.000}$\\
        Dance2Music \cite{lee2019dancing}                   & $0.097^{\pm.001}$ & $66.98^{\pm.016}$ & $8.116^{\pm.006}$ & $0.462^{\pm.011}$ & $0.043^{\pm.001}$\\
        TEMOS \cite{petrovich2022temos}                     & $0.722^{\pm.002}$ & $3.734^{\pm.028}$ & $3.703^{\pm.008}$ & $0.725^{\pm.071}$ & $0.368^{\pm.018}$\\
        TM2T \cite{guo2022tm2t}                             & $0.729^{\pm.002}$ & $1.501^{\pm.017}$ & $3.467^{\pm.011}$ & $8.973^{\pm.076}$ & $2.424^{\pm.093}$\\
        MLD \cite{chen2023executing}                        & $0.736^{\pm.002}$ & $1.087^{\pm.021}$ & $3.347^{\pm.008}$ & $8.589^{\pm.083}$ & $2.219^{\pm.074}$\\
        Guo \etal \cite{guo2022generating}                  & $0.772^{\pm.002}$ & $0.473^{\pm.013}$ & $3.196^{\pm.010}$ & $9.175^{\pm.082}$ & $2.413^{\pm.079}$\\
        MDM \cite{tevet2022human}                           & $0.611^{\pm.007}$ & $0.544^{\pm.044}$ & $5.566^{\pm.027}$ & $9.724^{\pm.086}$ & $2.799^{\pm.072}$\\
        MotionDiffuse \cite{zhang2022motiondiffuse}         & $0.782^{\pm.001}$ & $0.630^{\pm.001}$ & $3.113^{\pm.001}$ & $\textbf{9.410}^{\pm.049}$ & $1.553^{\pm.042}$\\
        T2M-GPT \cite{zhang2023generating}                  & $0.775^{\pm.002}$ & $0.116^{\pm.004}$ & $3.118^{\pm.011}$ & $9.761^{\pm.081}$ & $1.856^{\pm.011}$\\
        AttT2M \cite{zhong2023attt2m} & $0.786^{\pm.006}$ & $0.112^{\pm.006}$  & $3.038^{\pm.007}$ & $9.700^{\pm.090}$ & $\textbf{\bl{2.452}}^{\pm.051}$\\
        MAA  \cite{azadi2023make}                           & $0.675^{\pm.002}$ & $0.774^{\pm.007}$ & - & $8.230^{\pm.064}$ & -\\
        M2DM \cite{kong2023priority}                        & $0.763^{\pm.003}$ & $0.352^{\pm.005}$ & $3.134^{\pm.010}$ & $9.926^{\pm.073}$ & $\textbf{3.587}^{\pm.072}$\\ \hline
        \textbf{M2D2M (w/ $\beta_t$)}                                & $\textbf{\bl{0.796}}^{\pm.002}$ & $\textbf{\bl{0.115}}^{\pm.006}$ & $\textbf{\bl{3.036}}^{\pm.008}$ & $9.680^{\pm.074}$ & $2.193^{\pm.077}$\\
        \textbf{M2D2M (w/ $\beta(t,d)$)}                                & $\textbf{0.799}^{\pm.002}$ & $\textbf{0.087}^{\pm.004}$ & $\textbf{3.018}^{\pm.008}$ & $\textbf{\bl{9.672}}^{\pm.086}$ & $2.115^{\pm.079}$\\
        \hline  
    \end{tabular}
}
\label{table:single_hml3d_full}
\end{table}

\begin{table}[!t]
\centering
\caption{Single-motion generation performance on KIT-ML. The figures highlighted in \textbf{bold} and \textbf{\bl{blue}} denote the best and second-best results, respectively.}
\resizebox{.99\linewidth}{!}{
    \addtolength{\tabcolsep}{8pt}
    \begin{tabular}{l c c c c c}
        \hline
        Methods & R-Top3$\uparrow$ & FID$\downarrow$ & MM-Dist$\downarrow$ & Diversity$\rightarrow$ & MModality$\uparrow$ \\
        \hline\hline
        Ground Truth                                         & $0.779^{\pm.006}$ & $0.031^{\pm.004}$ & $2.788^{\pm.012}$ & $11.08^{\pm.097}$ & - \\
        VQ-VAE (reconstruction) & $0.740^{\pm.006}$ & $0.472^{\pm.011}$ & $2.986^{\pm.027}$ & $10.994^{\pm.120}$ & - \\
        \hline
        Seq2Seq \cite{lin2018generating}                    & $0.241^{\pm.006}$ & $24.86^{\pm.348}$ & $7.960^{\pm.031}$ & $6.744^{\pm.106}$ & - \\
        J2LP \cite{ahuja2019language2pose}                  & $0.483^{\pm.005}$ & $6.545^{\pm.072}$ & $5.147^{\pm.030}$ & $9.073^{\pm.100}$ & - \\
        Text2Gesture \cite{bhattacharya2021text2gestures}   & $0.338^{\pm.005}$ & $12.12^{\pm.183}$ & $6.964^{\pm.029}$ & $9.334^{\pm.079}$ & - \\
        Hier \cite{ghosh2021synthesis}                      & $0.531^{\pm.007}$ & $5.203^{\pm.107}$ & $4.986^{\pm.027}$ & $9.563^{\pm.072}$ & - \\
        MoCoGAN \cite{tulyakov2018mocogan}                  & $0.063^{\pm.003}$ & $82.69^{\pm.242}$ & $10.47^{\pm.012}$ & $3.091^{\pm.043}$ & $0.250^{\pm.009}$  \\
        Dance2Music \cite{lee2019dancing}                   & $0.086^{\pm.003}$ & $115.4^{\pm.240}$ & $10.40^{\pm.016}$ & $0.241^{\pm.004}$ & $0.062^{\pm.002}$  \\
        TEMOS \cite{petrovich2022temos}                     & $0.687^{\pm.002}$ & $3.717^{\pm.028}$ & $3.417^{\pm.008}$ & $10.84^{\pm.004}$ & $0.532^{\pm.018}$  \\
        TM2T \cite{guo2022tm2t}                             & $0.587^{\pm.005}$ & $3.599^{\pm.051}$ & $4.591^{\pm.019}$ & $9.473^{\pm.100}$ & $\textbf{\bl{3.292}}^{\pm.034}$  \\
        Guo \etal \cite{guo2022generating}                  & $0.681^{\pm.007}$ & $3.022^{\pm.107}$ & $3.488^{\pm.028}$ & $10.72^{\pm.145}$ & $2.052^{\pm.107}$  \\
        MLD \cite{chen2023executing}                        & $0.734^{\pm.007}$ & $\textbf{\bl{0.404}}^{\pm.027}$ & $3.204^{\pm.027}$ & $10.80^{\pm.117}$ & $2.192^{\pm.071}$  \\
        MDM \cite{tevet2022human}                           & $0.396^{\pm.004}$ & $0.497^{\pm.021}$ & $9.191^{\pm.022}$ & $10.847^{\pm.109}$ & $1.907^{\pm.214}$ \\
        MotionDiffuse \cite{zhang2022motiondiffuse}         & $0.739^{\pm.004}$ & $1.954^{\pm.062}$ & $\textbf{2.958}^{\pm.005}$ & $\textbf{11.10}^{\pm.143}$ & $0.730^{\pm.013}$  \\
        T2M-GPT \cite{zhang2023generating}                  & $0.737^{\pm.006}$ & $0.717^{\pm.041}$ & $3.053^{\pm.026}$ & $\textbf{\bl{10.862}}^{\pm.094}$ & $1.912^{\pm.036}$ \\
        AttT2M \cite{zhong2023attt2m} & $0.751^{\pm.006}$ & $0.870^{\pm.039}$ & $3.309^{\pm.021}$ & $10.96^{\pm.123}$ & $2.281^{\pm.047}$ \\
        M2DM \cite{kong2023priority}                         & $\textbf{\bl{0.743}}^{\pm.004}$ & $0.515^{\pm.029}$ & $3.015^{\pm.017}$ & $11.417^{\pm.097}$ & $\textbf{3.325}^{\pm037}$ \\\hline
        \textbf{M2D2M (w/ $\beta_t$)}                                & $\textbf{\bl{0.743}}^{\pm.006}$ & $\textbf{\bl{0.404}}^{\pm.022}$ & $3.018^{\pm.019}$ & ${10.749}^{\pm.102}$ & $2.063^{\pm.066}$ \\
        \textbf{M2D2M (w/ $\beta(t,d)$)}                                & $\textbf{0.753}^{\pm.006}$ & $\textbf{0.378}^{\pm.023}$ & $\textbf{\bl{3.012}}^{\pm.021}$ & ${10.709}^{\pm.121}$ & $2.061^{\pm.067}$ \\
        \hline
    \end{tabular}
}
\label{table:single_kitml_full}
\end{table}

In \Cref{table:single_hml3d_full,table:single_kitml_full}, we present a comprehensive comparison of single-motion generation results to demonstrate the effectiveness of our method.

\section{Analysis} \label{sec:analysis}
\subsection{Codebook visualization}
To examine relationships within the codebook, which inspired our design of dynamic transition probabilities as detailed in \cref{sec:transition}, we have visualized the tokens from the Motion VQ-VAE's codebook in \cref{fig:codebook}. This visualization reveals that certain tokens are more closely correlated, as evidenced by their clustering or alignment along implicit lines. Unlike the uniform transition strategy used in the VQ-Diffusion model, our method starts with a broad, exploratory range of transitions to encourage diversity by considering token proximity. These results justify our design of transition probabilities for the discrete diffusion.


\begin{figure}[t]
\centering
\includegraphics[width=0.55\linewidth]{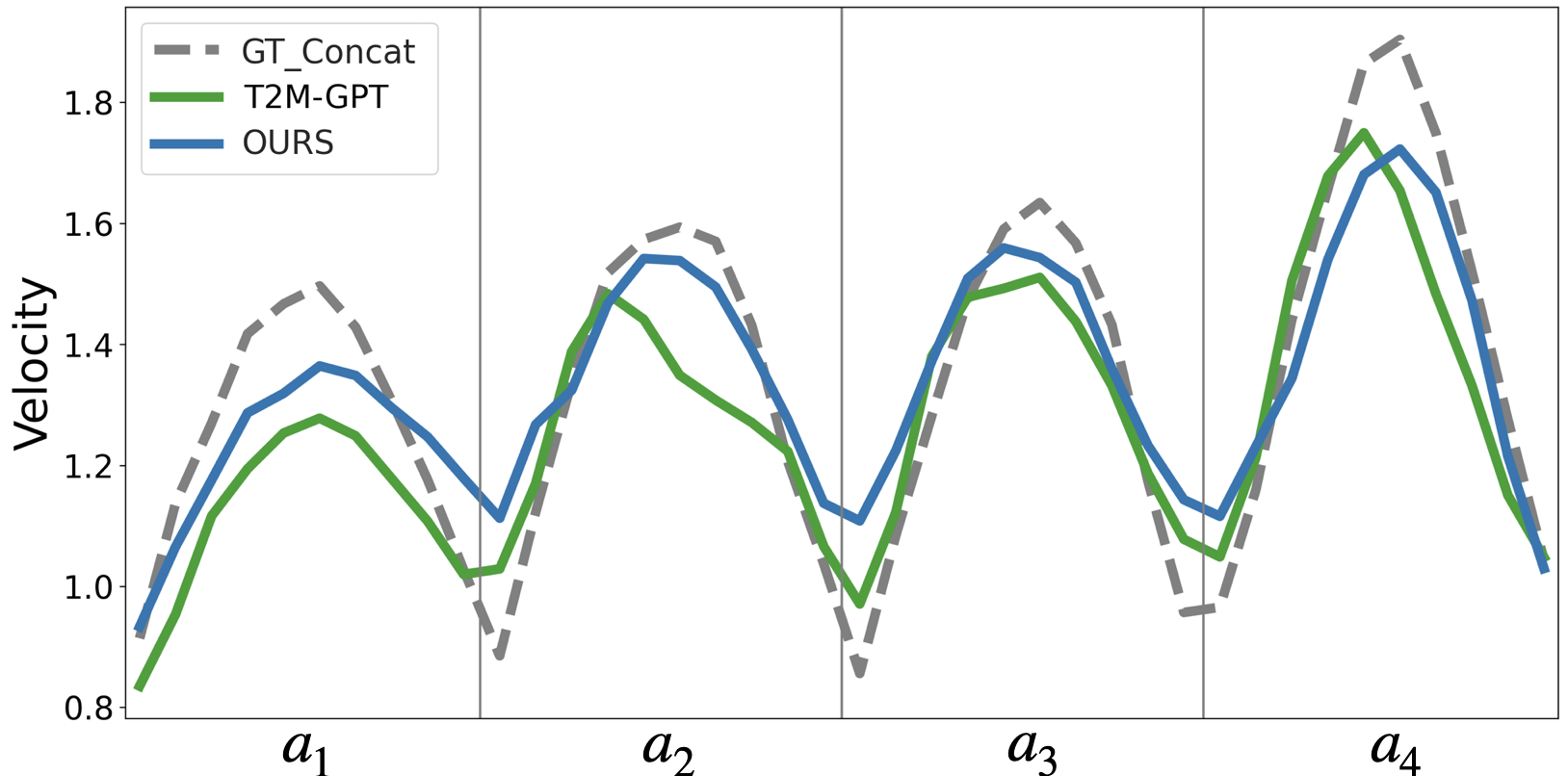}
\includegraphics[width=0.43\linewidth]{fig/transition.png}
\caption{$\textbf{(Left)}$ Plot of Mean Velocity and $\textbf{(Right)}$ plot of Mean Jerk of all transitions (40 frames) across all test sets in Multi-Motion Generation with  $N=4$.  `GT' represents concatenated real single motions.}
\label{fig:mean_velocity}
\end{figure}


\subsection{Mean Velocity \& Jerk Plot of Generated Multi-Motion}\vspace{-.5em}
To assess the smoothness of our M2D2M model, we plotted the mean velocity of the generated multi-motion sequences across all test sets for multi-motion generation, as shown in \cref{fig:mean_velocity}). In this figure, concatenated real single motions serve as the ground truth (GT). It is evident that the GT demonstrates discrete transitions between motions, while our M2D2M model (OURS) achieves smoother transitions with reduced jerk in the transitional phases.

\begin{figure}[!t]
\centering
\includegraphics[width=1.0\linewidth]{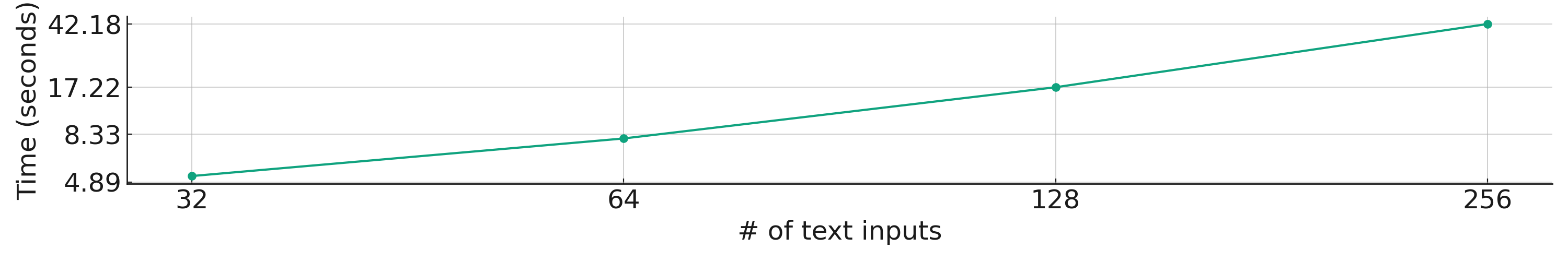}
\caption{Inference time scaling with action sequence length. Measured with a single NVIDIA RTX A6000 GPU.}
\label{fig:runtime}
\end{figure}

\subsection{Running time}
We calculate inference time based on the number of actions and visualize the results in \cref{fig:runtime}.
This illustration demonstrates that our method is practical for generating multi-motion sequences with reasonable computational cost.
We set each action to have 196 frames; thus, 256 text prompts generate 50,176 frames.
The gradient of the plotted line is nearly linear, as the joint sampling step is limited to $T_s$, allowing most other steps to be executed in parallel within a batch.

\section{Additional Qualitative Results of Generated Multi-Motion from M2D2M} \label{sec:addtional_qualitative}
Further qualitative results showcasing the capabilities of M2D2M in multi-motion generation, akin to the examples in \cref{fig:qualitative} in the main paper, are provided as animations (GIFs) in the supplementary materials.